\documentclass{article}
\pdfoutput=1

\usepackage[boxed, algoruled, vlined, linesnumbered, resetcount]{algorithm2e}

\usepackage{enumerate}
\usepackage{graphicx}
\usepackage[latin1]{inputenc}
\usepackage{palatino}
\usepackage{paralist}
\usepackage{latexsym}
\usepackage{listings} 
\usepackage{multicol}
\usepackage{natbib}
\usepackage{Sweave}
\usepackage{tabularx}
\usepackage{url}
\usepackage{xspace}

\newcommand{\vecx}{\ensuremath{\vec{x}}\xspace}

\newcommand{\bpm}{\begin{pmatrix}}
\newcommand{\epm}{\end{pmatrix}}

\newcommand{\bvm}{\begin{vmatrix}}
\newcommand{\evm}{\end{vmatrix}}

\newtheorem{definition}{Definition}[section]

\newcommand{\vecp}{\ensuremath{\vec{p}}\xspace}
\newcommand{\vecP}{\ensuremath{\vec{P}}\xspace}

\newcommand{\APD}{{\sc APD}\xspace}
\newcommand{\auto}{{\tt auto}\xspace}
\newcommand{\BST}{{\sc BST}\xspace}
\newcommand{\CONF}{{\sc CONF}\xspace}
\newcommand{\COUNT}{{\sc COUNT}\xspace}

\newcommand{\DE}{{\sc DE}\xspace}
\newcommand{\DES}{{\sc DES}\xspace}

\newcommand{\EDA}{{\sc EDA}\xspace}
\newcommand{\ES}{{\sc ES}\xspace}

\newcommand{\FACTOR}{{\tt FACTOR}\xspace}
\newcommand{\FLOAT}{{\tt FLOAT}\xspace}
\newcommand{\INT}{{\tt INT}\xspace}
\newcommand{\init}{{\tt init}\xspace}
\newcommand{\maxit}{ {\ensuremath{ \textrm{{\tt maxit}}}}\xspace}
\newcommand{\meta}{{\tt meta}\xspace}
\newcommand{\R}{{\sc R}\xspace}
\newcommand{\randomForest}{{\tt randomForest}\xspace}
\newcommand{\rep}{{\tt rep}\xspace}
\newcommand{\RES}{{\sc RES}\xspace}
\newcommand{\ROI}{{\sc ROI}\xspace}
\newcommand{\RSM}{{\sc RSM}\xspace}
\newcommand{\rsm}{{\tt rsm}\xspace}
\newcommand{\run}{{\tt run}\xspace}
\newcommand{\SANN}{{\sc SANN}\xspace}
\newcommand{\seq}{{\tt seq}\xspace}
\newcommand{\SPO}{{\sc SPO}\xspace}
\newcommand{\SPOT}{{\sc spot}\xspace}
\newcommand{\tmax}{{\tt tmax}\xspace}
\newcommand{\temp}{{\tt temp}\xspace}
\newcommand{\tree}{{\tt tree}\xspace}
\newcommand{\beq}{\begin{equation}}
\newcommand{\eeq}{\end{equation}}

\begin{document}

\title{\SPOT: An \R Package For Automatic and Interactive Tuning of
Optimization Algorithms\\ by Sequential Parameter Optimization}
\author{Thomas Bartz-Beielstein\\ Department of Computer Science,\\ 
Cologne University of Applied Sciences,\\
 51643 Gummersbach, Germany}

\maketitle

\begin{abstract}

The sequential parameter optimization (\SPOT) package for \R~\citep{R2008a} is a
toolbox for tuning and understanding simulation and optimization algorithms.
Model-based investigations are common approaches in simulation and
optimization. Sequential parameter optimization has been developed,
because there is a strong need for sound statistical analysis of simulation and
optimization algorithms.
\SPOT includes methods for tuning based on classical regression and analysis of
variance techniques; tree-based models such as CART and random forest; Gaussian
process models (Kriging), and combinations of different meta-modeling
approaches. This article exemplifies how \SPOT can be used for automatic and interactive
tuning.
\end{abstract}
%
%
%

\section{Introduction}
This article illustrates the functions of the \SPOT package. The \SPOT 
package can be downloaded from the comprehensive \R archive network at
\url{http://CRAN.R-project.org/package=SPOT}. \SPOT is
one possible implementation of the \emph{sequential parameter optimization}\/
(\SPO) framework introduced in~\citet{Bart06a}.
For a detailed documentation of the functions from the \SPOT package, the
reader is referred to the package help manuals.

The performance of modern search heuristics such as \emph{evolution
strategies}\,(\ES), \emph{differential evolution} (\DE), or \emph{simulated
annealing} (\SANN) relies crucially on their parametrizations---or, statistically speaking, on their factor settings. 
The term \emph{algorithm design} summarizes factors 
that influence the behavior (performance) of an algorithm, whereas \emph{problem design} 
refers to factors from the optimization (simulation) problem.
 Population size in \ES is one typical factor which belongs to the algorithm
 design, the search space dimension belongs to the problem design.
 We will consider \SANN in the remainder of this article, because it requires
 the specification of two algorithm parameters only.

One interesting goal of \SPO is to 
detect the importance of certain parts (subroutines such as recombination in
\ES) by systematically varying the factor settings of the algorithm design.
 This goal is related to improving the algorithm's \emph{efficiency}\/ and
 will be referred to in the following as \emph{algorithm tuning}, where the
 experimenter is seeking for an improved parameter setting, say $\vecp^*$, for
 one problem instance. Varying problem instances, e.g., search space dimensions or starting points of the algorithm, are associated with
 \emph{effectivity}\/ or the algorithm's robustness. In this case, the
 experimenter is interested in one parameter setting of the algorithm with which the algorithm performs
 sufficiently good on several problem instances. \SPOT can be applied
 for both tasks. The focus of this article lies on improving the algorithm's
 efficiency.
 
  Besides an improved performance of the algorithm, 
 \SPO  may lead to a better understanding of the algorithm.
\SPO combines several techniques from classical and modern statistics, 
namely \emph{design of experiments} (DoE) and 
\emph{design and analysis of computer experiments} (DACE)~\citep{Bart06a}. 
Basic ideas from \SPO rely heavily on Kleijnen's work on statistical
 techniques in simulation~\citep{Kle87a,Klei08a}.
 
Note, that we do not claim that \SPO is the only suitable way for tuning
algorithms. Far from it! We state that \SPO presents only one possible
way---which is possibly not the best for your specific problem. We highly recommend other
approaches in this field, namely F-race~\citep{Bira05a},
ParamILS~\citep{Hutt09e}, and REVAC~\citep{Nann09a}.

The paper is structured as follows:
Section~\ref{sec:motivation} presents an introductory example which
illustrates the use of tuning.
The sequential parameter optimization framework is presented in
Sect.~\ref{sec:spo}.
Details of the sequential parameter optimization toolbox are presented in
Sect.~\ref{sec:details}.
\SPOT uses plugins. Typical plugins are discussed in Sect.~\ref{sec:plugins}.
How \SPOT can be refined is exemplified in Sect.~\ref{sec:refine}.
Section~\ref{sec:summary} presents a summary and an outlook.

\section{Motivation}\label{sec:motivation}
\subsection{A Typical Situation}
We will discuss a typical situation from optimization. The practitioner is
interested in optimizing an objective function, say $f$,  with an
optimization algorithm $A$. She can use the optimization algorithm with default parameters.
This may lead to good results in some cases, whereas in other situations results are not
satisfactory. In the latter cases, practitioners try to determine improved
parameter settings for the algorithms manually, e.g., by changing one
algorithm parameter at a time. Before we will discuss problems related to this
approach, we will take a broader view and consider the general framework of
optimization via simulation which occurs in many real-world optimization scenarios.

\subsection{Optimization via Simulation}
\subsubsection{Modeling Layers}
\begin{figure}
\centering
\includegraphics[width=0.85\linewidth]{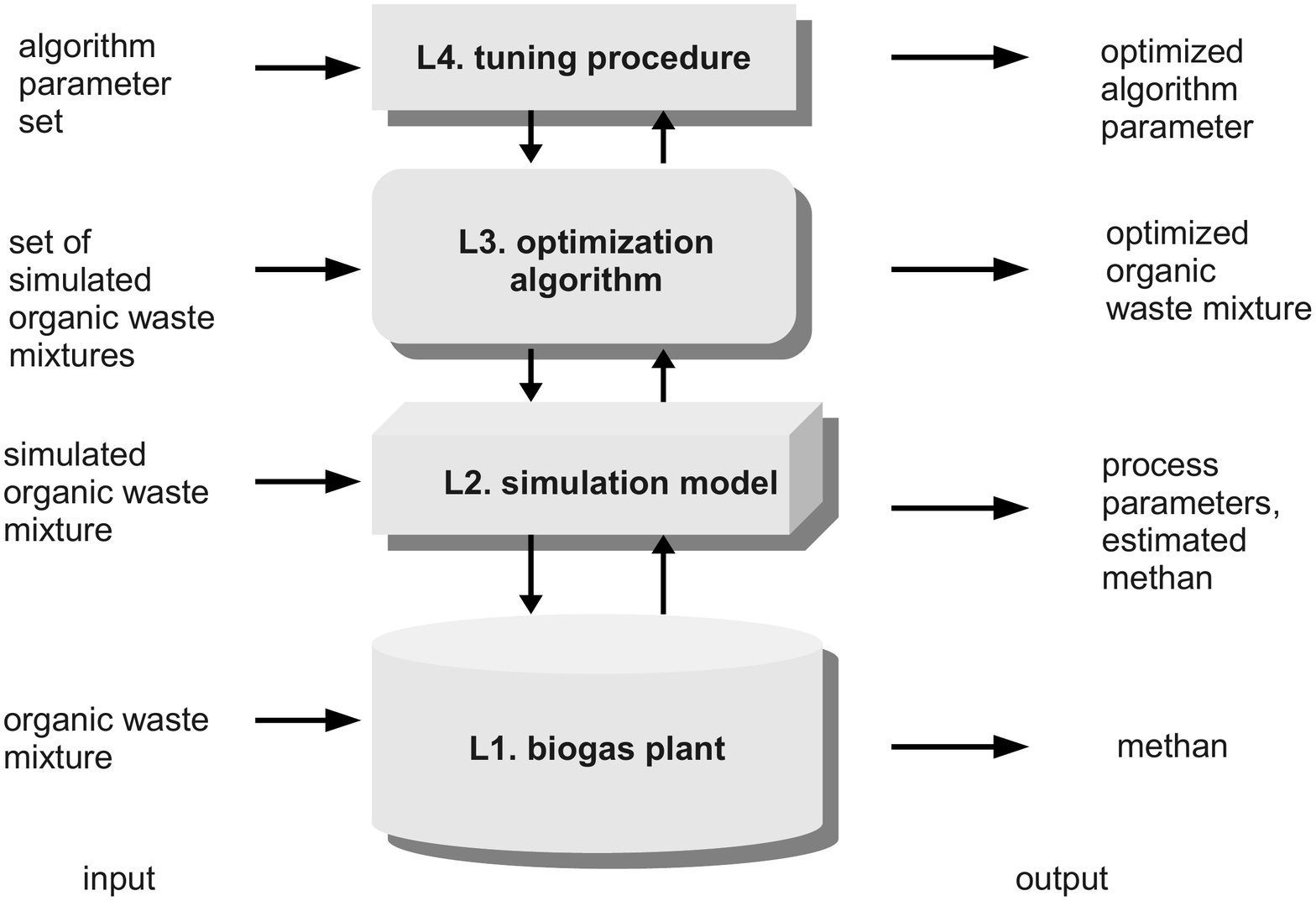}
\caption{Optimization via simulation. Illustration taken from~ \citet{Zieg10b},
who describe how \SPOT can be applied to the optimization of a
biogas-simulation model. Four different layers of the optimization of a biogas
simulation of shown. The first layer (L1) represents the real-world setting. Layer 2 (L2) shows the simulator. An objective function $f$ is defined at this layer. The
optimization algorithm $A$ belongs to the third layer (L3). The
fourth layer (L4) represents the algorithm tuning procedure, e.g., sequential parameter optimization.
\label{fig:biogassim}}
\end{figure}

To illustrate the task of optimization via simulation, the following layers can
be used.
\begin{description}
\item[(L1)] The real-world system, e.g., a biogas plant.
\item[(L2)] The related simulation model. The objective function $f$ is defined
at this layer. In optimization via simulation, problem parameters are defined at
this layer.
\item[(L3)] The optimization algorithm $A$. It requires the specification of
algorithm parameters, say  $\vecp^i \in \vecP$, where $\vecP$ denotes the set
of parameter vectors.
\item[(L4)] The experiments and the tuning procedure.
\end{description}
Figure~\ref{fig:biogassim} illustrates the situation.
To keep the setting as simple as possible, we consider an objective function
$f$ from layer (L2) and do not discuss interactions between (L1) and (L2).
Defining the relationship between (L1) and (L2) is not a trivial task. The
reader is referred to \citet{Law07a} and \citet{Fu02a} for an introduction.

\subsection{Description of the Objective Function}
\begin{figure}
\centering
\includegraphics[width=0.45\linewidth]{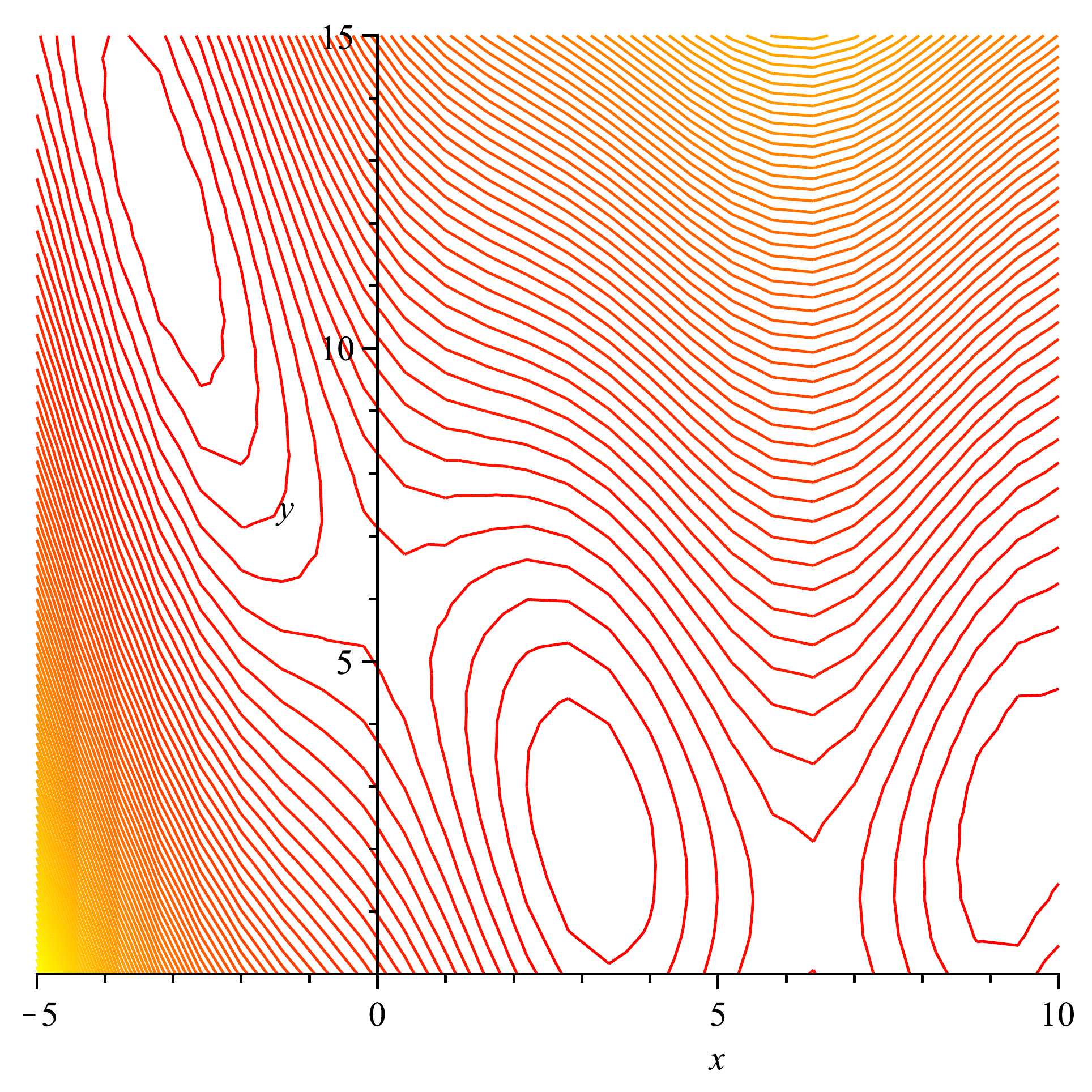}
\includegraphics[width=0.45\linewidth]{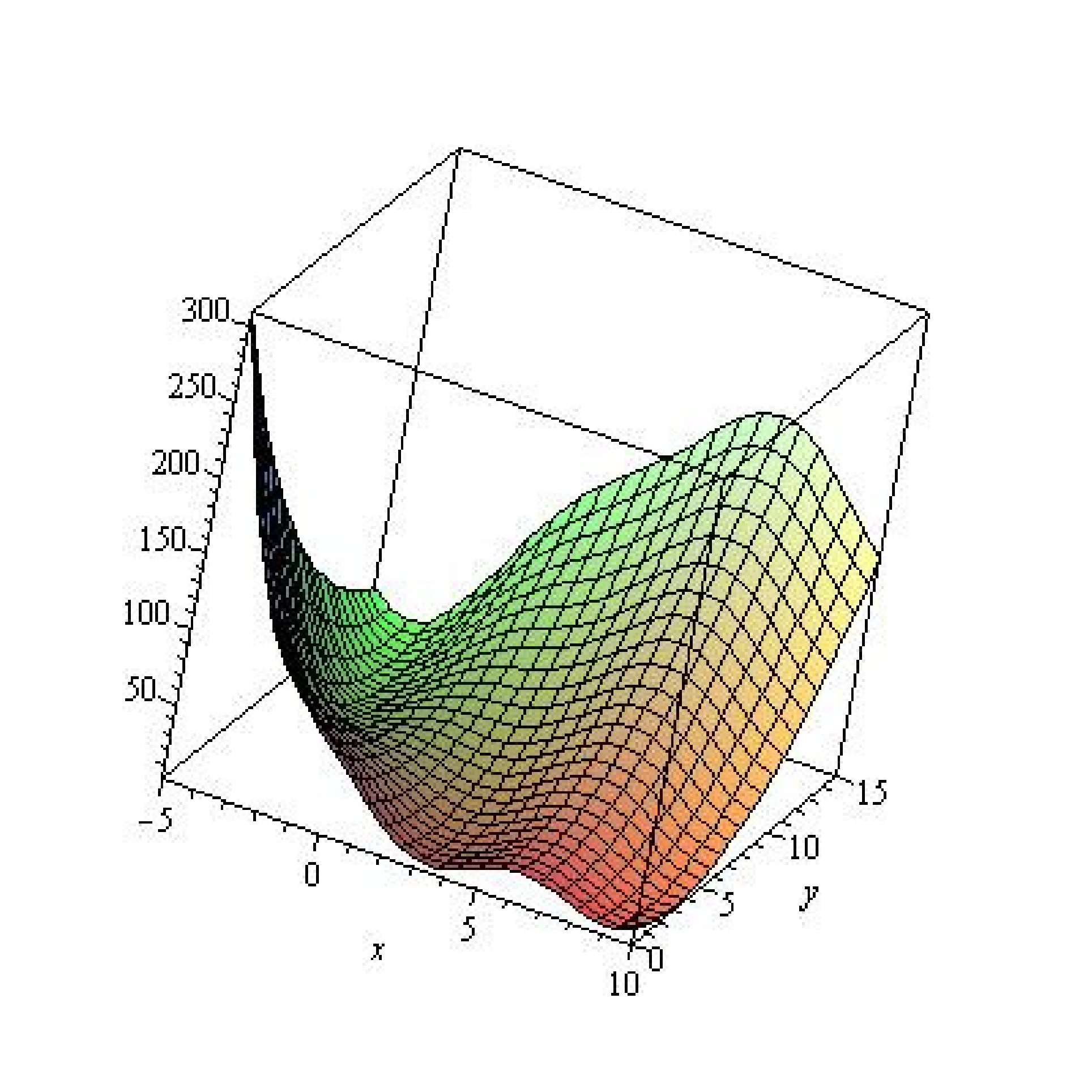}
\caption{Plots of the Branin function. The contour plot
(\emph{left}) shows the location of the three global
minima\label{fig:braninPlot}}
\end{figure}

The Branin function
\[
f(x_1,x_2) = \left(x_2 - \frac{5.1}{4\pi^2} x_1^2 + \frac{5}{\pi} x_1 -6
\right)^2 + 10 \times \left(1-\frac{1}{8\pi}\right) \cos(x_1) +10,
\]
with
\beq\label{eq:roi0005a}
x_1\in [-5,10] \textrm{ and }  x_2 \in [0,15].
\eeq
was chosen as a test function,
because it is well-known in the global optimization community, so results are
comparable.
It has three global minima, $\vecx_1^*=[3.1416, 2.2750]$,
$\vecx_2^*=[9.4248, 2.4750]$ and
$\vecx_3^*=[-3.1416, 12.2750]$ with
$y^* = f(\vecx_i^*)= 0.39789$, ($i=1,2,3$), see Fig.~\ref{fig:braninPlot}.

\subsection{Description of the Optimization Algorithm}
In order to improve reproducibility of the examples presented in this article,
an algorithm which is an integral part of the \R system, the method \SANN,
was chosen. It is described in \R's help system as follows~\citep{R2008a}: 
Method \SANN is by default a variant of simulated annealing given in~\citet{Beli92a}. 
Simulated annealing belongs to the class of stochastic global optimization
methods. It uses only function values but is relatively slow. It will also work for non-differentiable functions. This implementation uses the Metropolis function for the acceptance probability.
By default the next candidate point is generated from a
Gaussian Markov kernel with scale proportional to the actual temperature.
If a function to generate a new candidate point is given, method \SANN can
also be used to solve combinatorial optimization problems. Temperatures are decreased according to the logarithmic cooling schedule as
given in~\citet{Beli92a}; specifically, the temperature is set to
\begin{Schunk}
\begin{Sinput}
temp / log(((t-1) \%/\% tmax)*tmax + exp(1))
\end{Sinput}
\end{Schunk}
 where $t$ is the current iteration step and \temp and \tmax are specifiable via
 control. Note that the \SANN method depends critically on the settings of
 the control parameters. Summarizing, there are two algorithm parameters which have to be specified
before the algorithm is run:
 \begin{enumerate}
\item \temp controls the \SANN method. It is the starting temperature for
the cooling schedule. Defaults to 10.
\item \tmax is the number of function evaluations at each temperature for the
\SANN method. Defaults to 10.
 \end{enumerate}
 Note, \tmax is an integer. How different parameter types can be handled is
 described in Sec.~\ref{sec:cate}. To simply the discussion, \temp will be
 treated as a numerical value in the remainder of this article.
 
\subsection{Starting Optimization Runs}\label{sec:starting}
Now we discuss the typical situation from optimization:  An experimenter 
applies an optimization algorithm $A$ (\SANN) to an 
objective function $f$ (Branin function) in order to determine the minimum.

First, we will set the seed to obtain reproducible results. 
\begin{Schunk}
\begin{Sinput}
> set.seed(1)
\end{Sinput}
\end{Schunk}
Next, we will define the objective function.
\begin{Schunk}
\begin{Sinput}
> spotFunctionBranin <- function(x) {
+     x1 <- x[1]
+     x2 <- x[2]
+     (x2 - 5.1/(4 * pi^2) * (x1^2) + 5/pi * x1 - 6)^2 + 10 * (1 - 
+         1/(8 * pi)) * cos(x1) + 10
+ }
\end{Sinput}
\end{Schunk}
Then, the starting point for the optimization, $\vecx_0$,
 and the number of function evaluations, $\maxit$, are defined:
\begin{Schunk}
\begin{Sinput}
> x0 <- c(10, 10)
> maxit <- 250
\end{Sinput}
\end{Schunk}
The parameters specified so far belong to the problem design. 
Now we have to consider parameters from the algorithm design, i.e., parameters that 
control the behavior of the \SANN algorithm, namely \tmax and \temp. 
Default values are chosen first:
\begin{Schunk}
\begin{Sinput}
> tmax <- 10
> temp <- 10
\end{Sinput}
\end{Schunk}
Finally, we can start the optimization algorithm (\SANN):
\begin{Schunk}
\begin{Sinput}
> y1 <- optim(x0, spotFunctionBranin, method = "SANN", control = 
+     list(maxit = maxit, temp = temp, tmax = tmax))
\end{Sinput}
\end{Schunk}
\SANN returns the following result:
\begin{Schunk}
\begin{Sinput}
> print(y1$value)
\end{Sinput}
\begin{Soutput}
[1] 4.067359
\end{Soutput}
\end{Schunk}
Since the optimum value reads $y^* = 0.39789$, the practitioner is interested in improving this result by modifying the 
algorithm parameters \tmax and \temp:
\begin{Schunk}
\begin{Sinput}
> tmax <- 10
> temp <- 20
> y2 <- optim(x0, spotFunctionBranin, method = "SANN",
+    control = list(maxit = maxit, temp = temp, tmax = tmax))
\end{Sinput}
\end{Schunk}
Results obtained with the new \tmax and \temp values look promising:
\begin{Schunk}
\begin{Sinput}
> print(y2$value)
\end{Sinput}
\begin{Soutput}
[1] 0.4570975
\end{Soutput}
\end{Schunk}
However, since \SANN is a stochastic algorithm, the practitioner wants to investigate
the dependency of the results on the random seed. So she performs the same experiment
with modified seed.
\begin{Schunk}
\begin{Sinput}
> set.seed(1000)
> y3 <- optim(x0, spotFunctionBranin, method = "SANN", 
+       control = list(maxit = maxit, temp = temp, tmax = tmax))
> print(y3$value)
\end{Sinput}
\begin{Soutput}
[1] 7.989125
\end{Soutput}
\end{Schunk}
This result is rather disappointing, because a worse value is obtained with this 
seemingly better parameter settings.
%
Results from these experiments are summarized in Tab.~\ref{tab:sum1}.
\begin{table}
\caption{Results from manually tuning \SANN on Branin function.~\label{tab:sum1}
Smaller values are better. Run 1 reports results from the default configuration.
Run 2 uses a different temperature and obtains a better function value. 
However, this result cannot be generalized, because modifying the seed
leads to a worse function value}
\begin{tabularx}{\textwidth}{XXXXX}
run & \temp & \tmax & seed & result\\
\hline
1 & 10 & 10 & 1 & 4.067359\\
2 & 20 & 10 & 1 & 0.4570975\\
3 & 20 & 10 & 1000 & 7.989125
\end{tabularx}
\end{table}

The practitioner has modified one variable (\temp) only. Introducing
variations of the second variable (\tmax) complicates the situation, because
interactions between these two variables might occur. And, the experimenter has
to take random effects into account.
Here comes \SPOT into play.
\SPOT was developed for tuning algorithms in a reproducible way. It uses
results from algorithm runs to build up a meta model. This meta model enables
the experimenter to detect important input variables, estimate effects, and
determine improved algorithm configurations in a reproducible manner. Last but
not least, the experimenter learns from these results.

One simple goal, which can be tackled with \SPOT, is to determine the best
parameter setting of the optimization algorithm for one specific instance of an
optimization problem.
It is not easy to define the term ``best'', because it can be defined in many ways
and this definition is usually problem specific. \citet{Klei02b} presents
interesting aspects from practice. See also the discussion
in chapter 7 of~\citet{Bart06a}. Therefore, we will take a
naive approach by defining our tuning goal as the following hypothesis:
\begin{description}
\item[(H-1)] ``We can determine a parameter setting $\vecp^*$ which improves
\SANN's performance. To measure this performance gain, the
average function values from ten runs of \SANN with default, i.e., $\vecp^0$
and tuned parameter $\vecp^*$ settings are compared.''
\end{description}

\subsection{Tuning with \SPOT }\label{sec:tunespot}
Before \SPOT is described in detail, we will demonstrate how it can be applied
to find an answer for hypothesis (H-1).

\subsubsection{\SPOT Projects}
A \SPOT \emph{project}\/ consists of a set of files with the same basename, but
different extensions, namely \CONF, \ROI, and \APD. Here, we will discuss the
project {\tt demo07RandomForestSann}, which is included in the \SPOT package,
see
\begin{Schunk}
\begin{Sinput}
> demo(package="SPOT")
\end{Sinput}
\end{Schunk}
for demos in the \SPOT package.
Demo projects, which are included in the \SPOT package, can be found in the
directory of your local \SPOT installation, e.g.,
\url{~/Ri486-pc-linux-gnu-library/2.11/SPOT} on Linux systems.

A \emph{configuration}\/ (\CONF) file, which stores information about \SPOT
specific settings, has to be set up. 
For example, the number of \SANN algorithm runs, i.e., the available budget, can
be specified via {\tt auto.loop.nevals}.
\SPOT implements a sequential approach, i.e., the available budget is not used
in one step. Evaluations of the algorithm on a subset of this budget, the
so-called initial design, is used to generate a coarse grained meta model $F$.
This meta model is used to determine promising algorithm design points
which will be evaluated next. 
Results from these additional \SANN runs are used to refine the meta model $F$.
The size of the initial design can be specified via {\tt init.design.size}.
To generate the meta model, we use random forest~\citep{Brei01a}. This can be
specified via {\tt seq.predictionModel.func = "spotPredictRandomForest"}.
Available meta models are listed in Sect.~\ref{sec:seq}. 
Random forest was chosen, because it is a robust method which can handle
categorical and numerical variables.
In the following
example, we will use the configuration file {\tt demo07RandomForestSann.conf}.

A \emph{region of interest}\/ (\ROI) file specifies algorithm parameters and
associated lower and upper bounds for the algorithm parameters.  Values for \temp are chosen from
the interval $[1;50]$. 
{\tt TEMP 1 50 FLOAT} is the corresponding line for the \temp parameter which
is added to the file {\tt demo07RandomForestSann.roi}.

Optionally, an \emph{algorithm problem design}\/ (APD) file can be
specified. This file contains information about the problem and might be used by the algorithm. For
example, the starting point {\tt x0 = c(10,10)} can be specified in the \APD
file. The file {\tt demo07RandomForestSann.apd} will be used in our example.

\subsubsection{Starting \SPOT in Automatic Mode}\label{sec:auto}
\begin{figure}
\centering
\includegraphics[width=\linewidth]{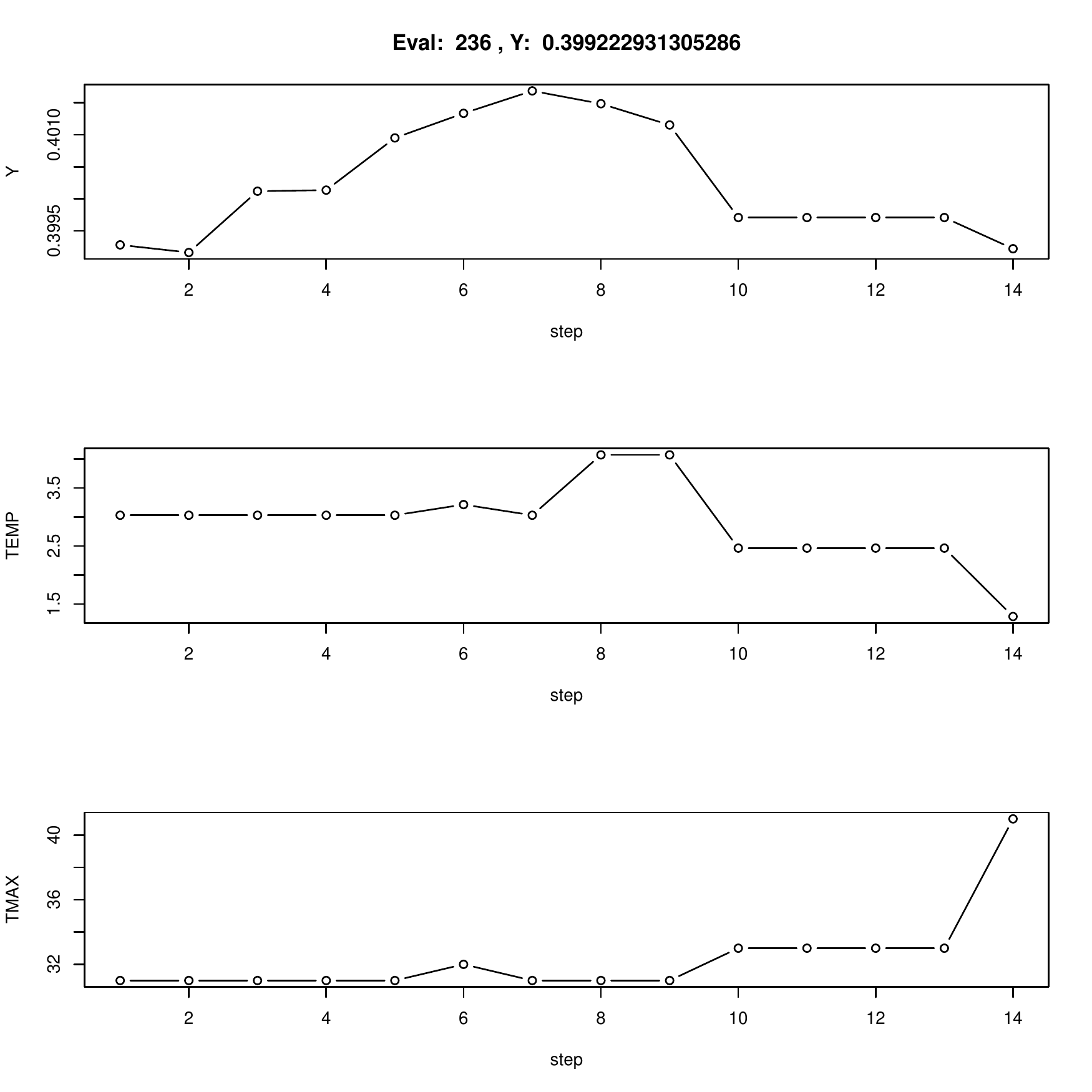}
\caption{Tuning \SANN with \SPOT. Random forest was chosen as a meta model.
This output can also be shown on-line, i.e., during the tuning process in order
to visualize the progress. The first panel shows the average
function value of the best configuration found so far. The second and third panel 
visualize corresponding parameter settings. These values
are updated each time a new meta model 
(random forest) is build. Each time a meta model is build, the step counter is
increased. Altogether 14 meta models (random forest) are build during this
tuning process and 236 runs of the \SANN algorithm were executed}
\label{fig:final}
\end{figure}

\begin{figure}
\centering
\includegraphics[width=\linewidth]{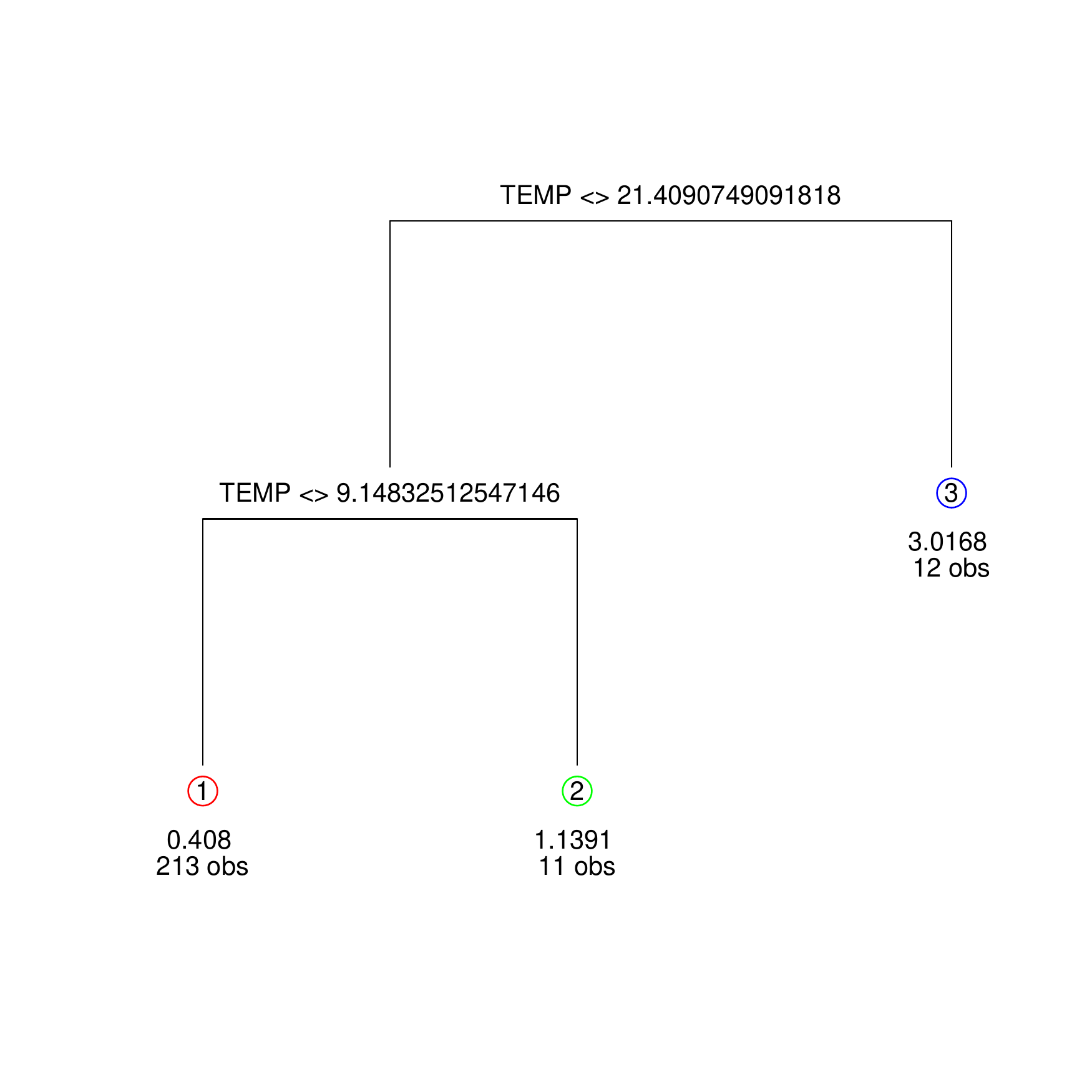}
\caption{Tuning \SANN with \SPOT. Random forest was chosen as a meta model. This
simple regression tree is generated by \SPOT's default report function {\tt
spotReportDefault}. The tree illustrates that \temp has the largest effect.
Values at the terminal node $t_i$ show the average function value and the number
of observations (obs) which fulfill the conditions which are given by following
the tree from the root node to $t_i$. Smaller \temp values improve
\SANN's performance. A value of \temp, which is smaller than 9.14, results in an average function value of $\overline{y}=0.408$. This result is based on 213 observations}
\label{fig:tune1}
\end{figure}

If these files are available, \SPOT can be started from \R's command line via
\begin{Schunk}
\begin{Sinput}
> library(SPOT)
> spot(``demo07RandomForestSann.conf'')
\end{Sinput}
\end{Schunk}
\SPOT is run in automatic mode, if no task is specified (this is the default
setting). Result from this run reads
\begin{Schunk}
\begin{Soutput}
Best solution found with 236 evaluations:
    Y     TEMP     TMAX   COUNT CONFIG
0.3992229 1.283295   41    10     36
\end{Soutput}
\end{Schunk}
\SPOT has determined a configuration $\temp = 1.283295$ and $\tmax =
41$, which gives an average function value from ten runs of $\overline{y}=
0.3998429$. \SPOT uses an internal counter (\COUNT) for configurations. The best
solution was found with configuration 36.
The tuning process is illustrated in
Fig.~\ref{fig:final}. Figure~\ref{fig:tune1} shows a regression tree which is
generated by the default report function {\tt spotReportDefault}.

\subsection{Validating the Results}
Finally, we will evaluate the result by comparing ten runs of \SANN with default
parameter settings, say $\vecp^0$ to ten runs with the tuned configurations from
\SPOT, say $\vecp^*$. The corresponding \R commands used for this comparison are
shown in the Appendix.
First, we will set the seed to obtain reproducible results. 
Next, we will define the objective function.
Then, the starting point for the optimization $ \vecx_0 $ and 
the number of function evaluations \maxit are defined.
The parameters specified so far belong to the problem design.
 
Now we have to consider parameters from the algorithm design, i.e., parameters that 
control the behavior of the \SANN algorithm, namely \tmax and \temp.
Finally, we can start the optimization algorithm (\SANN).
The run is finished with the following summary:
\begin{Schunk}
\begin{Soutput}
   Min. 1st Qu.  Median    Mean 3rd Qu.    Max. 
 0.3995  0.4037  0.4174  0.9716  0.6577  4.0670 
\end{Soutput}
\end{Schunk}

In order to illustrate the performance gain from \SPOT's tuning procedure,  \SANN is run
 with the tuned parameter configuration, i.e.,
$\temp = 1.283295$ and $\tmax = 41$. Results from these ten \SANN runs can be
summarized as follows:
 \begin{Schunk}
   \begin{Soutput}
   Min. 1st Qu.  Median    Mean 3rd Qu.    Max. 
 0.3980  0.3982  0.3984  0.3995  0.3989  0.4047 
\end{Soutput}
\end{Schunk}

\begin{figure}
\begin{center}
\includegraphics[width=0.5\linewidth]{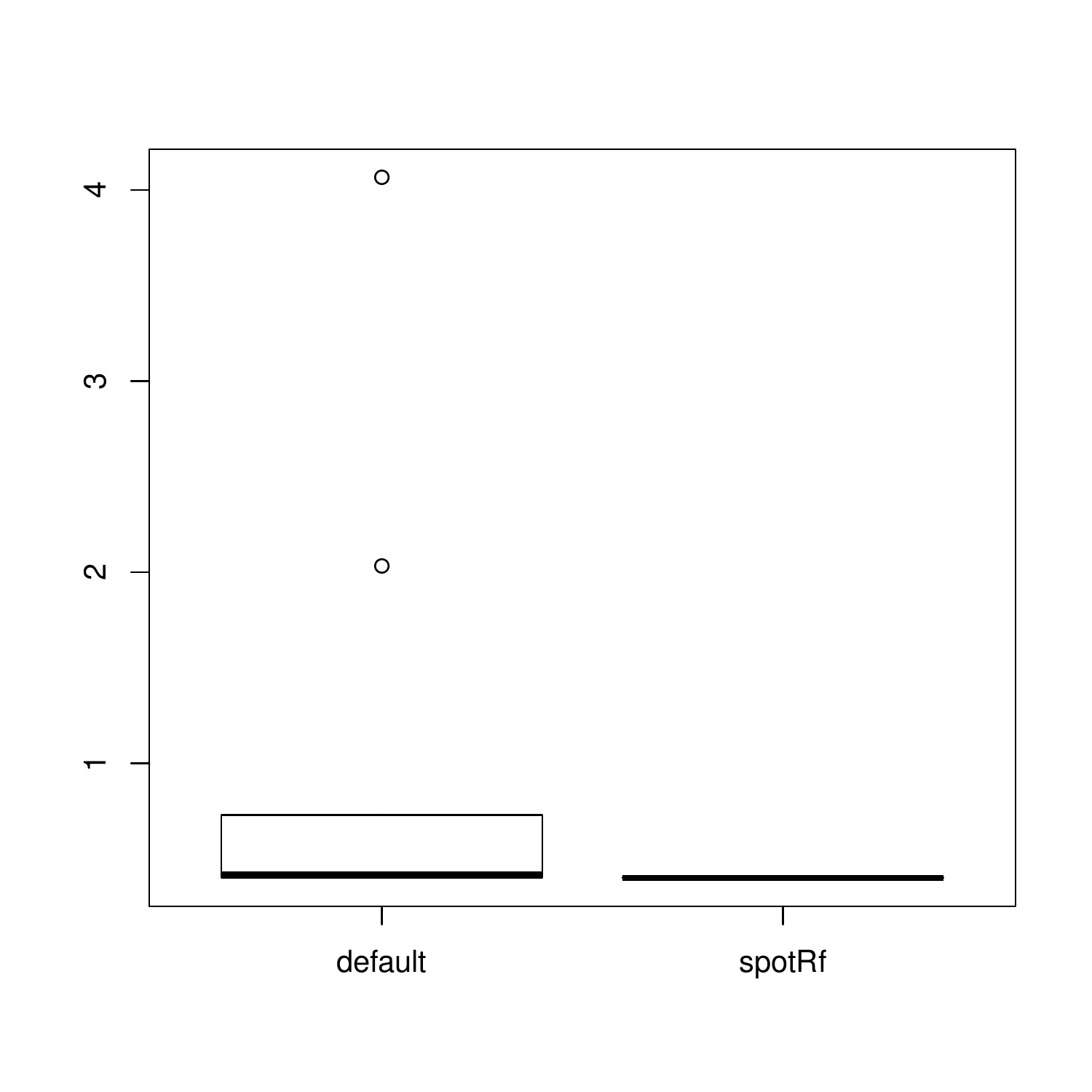}
\end{center}
\caption{Comparison of \SANN's default parameter values (\emph{default}) with
parameter settings obtained with \SPOT, where random forest was chosen as a
meta model (\emph{spotRf}) 
\label{fig:box1}}
\end{figure}

Further statistical analyses, e.g., the box plot shown in
Fig.~\ref{fig:box1}, reveal that this difference is statistically
significant.
Hence, hypothesis (H-1) cannot be rejected.
After this quick introduction we will have a closer look at \SPOT.

\section{Sequential Parameter Optimization}\label{sec:spo}
\subsection{Definition}
\begin{definition}[Sequential Parameter Optimization]\label{def:spo}
\emph{Sequential parameter optimization}\/
(\SPO) is a framework for tuning and understanding of algorithms by active
experimentation.
\SPO employs methods from error statistics to obtain reliable results.
It comprises the following elements:
\begin{compactenum}[SPO-1: ]
\begin{multicols}{2}
\item Scientific questions
\item Statistical hypotheses
\item Experiments
\item Scientific meaning
\end{multicols}
\end{compactenum}
\hfill $\Box$
\end{definition}

\begin{figure}
\includegraphics[width=\textwidth]{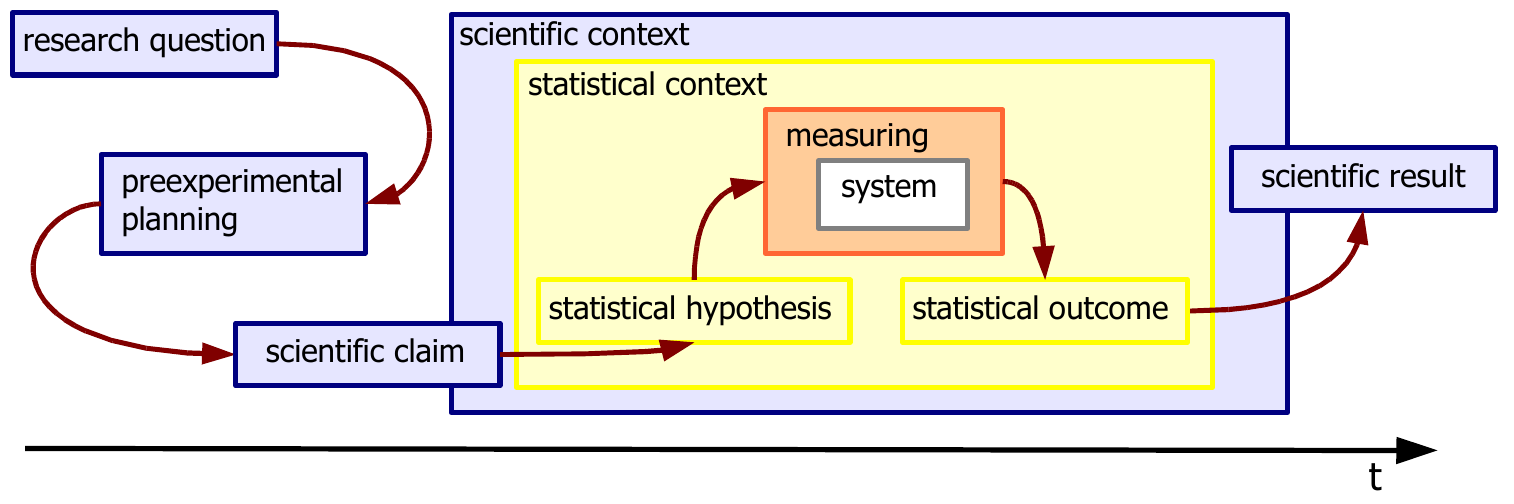}
\caption{Steps and contexts of performing an experiment from research question
to scientific result}
\label{fig:contexts}
\end{figure}

These elements can be explained as follows.
Starting point of the investigation is a scientific question (SPO-1).
This question often deals with assumptions about algorithms, e.g., influence of parameter values or new
operators.
This (complex) question is broken down into (simple) statistical
hypotheses (SPO-2) for testing, see~\cite{Bart08a} for an example.
Next, experiments can be performed for each hypothesis, e.g., (H-1) as defined
in Sect.~\ref{sec:starting}.
 \begin{enumerate}[a) ]
  \item Select a model $F$ (e.g., random forest) to describe
  a functional relationship\label{enu:step1}.
  \item Select an experimental design, e.g., Latin hypercube design.
  \item Generate data, i.e., perform experiments.
  \item Refine the model until the hypothesis can be
  accepted/rejected\label{enu:step2}.
 \end{enumerate}
Performing these experiments will be referred to as step (SPO-3).
Finally, to assess the scientific meaning of
the results from an experiment, conclusions are drawn from the hypotheses.
This is step (SPO-4) in the sequential parameter optimization framework, see
Definition~\ref{def:spo}.
Figure~\ref{fig:contexts} illustrates the \SPO framework. \SPOT implements the
steps from the statistical context.

This article describes one specific instance of (SPO-3), which implements the
corresponding software programs in \R. It will be referred to as \SPOT.


\subsection{Sequential Parameter Optimization
Toolbox}\label{sec:spot}
We introduce \R's \SPOT package as one possible implementation of step (\SPO-3)
from the \SPO framework. 
Implementations in other programming languages, e.g., MATLAB, are also
available but are not subject of this article.

The \SPO \emph{toolbox} was developed over recent years by Thomas
Bartz-Beielstein, Christian Lasarczyk, and Mike Preuss~\citep{BLP05}. Main
goals of \SPOT are (i) the  determination of  improved parameter settings for
optimization algorithms and (ii) to provide statistical tools for analyzing  and
understanding their performance.

\begin{definition}[Sequential Parameter Optimization
Toolbox]\label{def:spot}\label{sec:practical}
The sequential parameter optimization toolbox implements the following
features, which are related to step (SPO-3) from the \SPO framework.
\begin{compactenum}[SPOT-1: ]
\item Use the available budget (e.g., simulator runs, number of
function evaluations) sequentially, i.e., use information from the
exploration of the search space to guide the search by building one or several meta models. Choose new design points based on predictions from the
meta model(s). Refine the meta model(s) stepwise to improve knowledge about
the search space.
\item If necessary, try to cope with noise  by
improving confidence. Guarantee comparable confidence for search points.
\item Collect information to learn from this tuning process, e.g., apply
exploratory data analysis.
\item Provide mechanisms both for interactive and automatic tuning.
\end{compactenum}
\hfill $\Box$
\end{definition}
The article entitled ``sequential parameter
optimization"~\citep{BLP05} was the first attempt to
summarize results from seminars and tutorials given at conferences such as CEC
and GECCO and make this approach known to and available for a broader
audience~\citep{Bei02lec,BP04lec,Bar05lec,BP05blec,BP05alec}.

\SPOT was successfully applied in the fields of
bioinformatics~\citep{Volk06a, Fobe09a}, environmental
engineering~\citep{Kone09b,Flas10a}, shipbuilding~\citep{Rudo09c}, fuzzy
logic~\citep{Yi08a}, multimodal optimization~\citep{Preu07b}, statistical
analysis of algorithms~\citep{Lasa07a, Trau09a}, multicriteria
optimization~\citep{Bart09l}, genetic programming~\citep{Lasa05a},
particle swarm optimization~\citep{BPV04b,Kram07a},
automatic and manual parameter tuning~\citep{Fobe06a,Smit09a,Hutt09a,Hutt10a},
graph drawing~\citep{Tosi06a, Poth07a},
aerospace and shipbuilding industry~\citep{Nauj06a},
mechanical engineering~\citep{MMLB07a},
and
chemical engineering~\citep{Henr08a}.
\citet{Bart10b}  collects more than 100 publications related to
the sequential parameter optimization.

\subsection{Elements of the SPOT Framework}\label{sec:spotelements}
\subsubsection{The General SPOT Scheme}\label{sec:spo-algorithm}
\begin{algorithm}[tb]
\begin{small}
\caption{SPOT\label{fig:pseudo-spo}}
\label{alg:spo-algorithm}
\tcp{phase 1, building the model:}
let $A$ be the tuned algorithm\;\nllabel{spo-p1-start}
generate an initial population $\vecP = \{\vecp^1, \dots, \vecp^m\}$
of $m$ parameter vectors\nllabel{spo-alg:initial}\; let $k = k_0$ be the
initial number of tests for determining estimated utilities\;
\ForEach{$\vecp^i \in \vecP$}{
\nllabel{spo:loop}
run $A$ with $\vecp^i$ $k$ times to determine the estimated utility $u^i$
(e.g., average function value from 10 runs) of $\vecp^i$\;\nllabel{spo-p1-end}
}
\tcp{phase 2, using and improving the model:}
\While{termination criterion not true}{
let $\vecp^*$ denote the parameter vector from $\vecP$ with best estimated
utility\; let $k$ the number of repeats already computed for $\vecp^*$\;
build prediction model $F$ based on $\vecP$ and $u^1, \dots,
u^{|\vecP|}\}$\; generate a set $\vecP'$ of $l$ new parameter vectors by
random sampling\; \ForEach{$\vecp^i \in \vecP'$}{
calculate $f(\vecp^i)$ to determine the predicted utility $F(\vecp^i)$\;
}
\nllabel{spo-alg:d}
select set $\vecP''$ of $d$ parameter vectors from $\vecP'$ with best
predicted utility ($d \ll l$)\; run $A$ with $\vecp^*$ once and recalculate
its estimated utility using all $k + 1$ test results; \tcp{(improve
confidence)} update $k$, e.g., let $k = k + 1$\; run $A$ $k$ times with each
$\vecp^i \in \vecP''$ to determine the estimated utility $F(\vecp^i)$\; extend
the population by $\vecP = \vecP \cup \vecP''$\;\nllabel{spo-p2-end}
}\nllabel{spo-p2-start}
\end{small}
\end{algorithm}

Algorithm~\ref{alg:spo-algorithm} presents a formal description of the \SPOT
scheme. 
The \emph{utility} is used to measure algorithm's performance. Typical measures
are the estimated mean or median from several runs of $A$.
Algorithm~\ref{alg:spo-algorithm} consists of two phases, namely the first construction of the model (lines \ref{spo-p1-start}--\ref{spo-p1-end}) and
its sequential improvement (lines \ref{spo-p2-start}--\ref{spo-p2-end}). Phase 1
determines a population of initial designs in algorithm parameter space and runs
the algorithm $k$ times for each design. Phase 2 consists of a loop with the
following components:
\begin{enumerate}
  \item Update the meta model $F$ (or several meta models $F_i$) by means of the
  obtained data.
  \item Generate a (large) set of design points and compute their utility by
  sampling the model.
  \item  Select the seemingly best design points and run the algorithm for
  these.
  \item  The new design points are added to the population
  and the loop starts over if the termination criterion is not reached.
\end{enumerate}
A counter $k$ is increased in each cycle and used to determine the
number of repeats that are performed for each setting to be statistically sound in the
obtained results. In consequence, this means that the best design points so far
are also run again to obtain a comparable number of repeats. These
reevaluations may worsen the estimated performance and explains increasing
$Y$ values in Fig.~\ref{fig:final}.

Sequential approaches can be more efficient than approaches that evaluate the
information in one step only~\citep{Wald47a}. This presumes an experienced
operator who is able to draw the right conclusions out of the first results. In case the operator is new to \SPOT  the sequential steps can be started automatically. 
Compared to interactive procedures, performance in the automatic tuning
process may decrease. However, results from different algorithm runs, e.g.,
\ES and \SANN, will be comparable in an objective manner if data for the
comparison is based on the same tuning procedure.

Extensions to the \SPOT approach were proposed
by other authors, e.g., \citet{Lasa07a} integrated an \emph{optimal computational budget allocation}  procedure, which is based on ideas
by \citet{Chen05a}. Due to \SPOT's plugin structure, see
Sect.~\ref{sec:plugins}, further extensions can easily be integrated.

\subsubsection{Running SPOT}
In Sect.~\ref{sec:auto}, \SPOT was run as an automatic tuner. Steps from the
automatic mode can be used in an interactive manner.
 \SPOT can be started with the command
\begin{verbatim}
spot (<configurationfile>,  <task>)
\end{verbatim}
where 
{\tt configurationfile} is the name of the \SPOT configuration file and
{\tt task} can be one of the tasks  \init, \seq, \run, \rep or \auto.
\SPOT can also be run in a \meta mode to perform tuning over a set of problem
instances. 

\begin{figure}
\begin{center}
\includegraphics[width=0.85\linewidth]{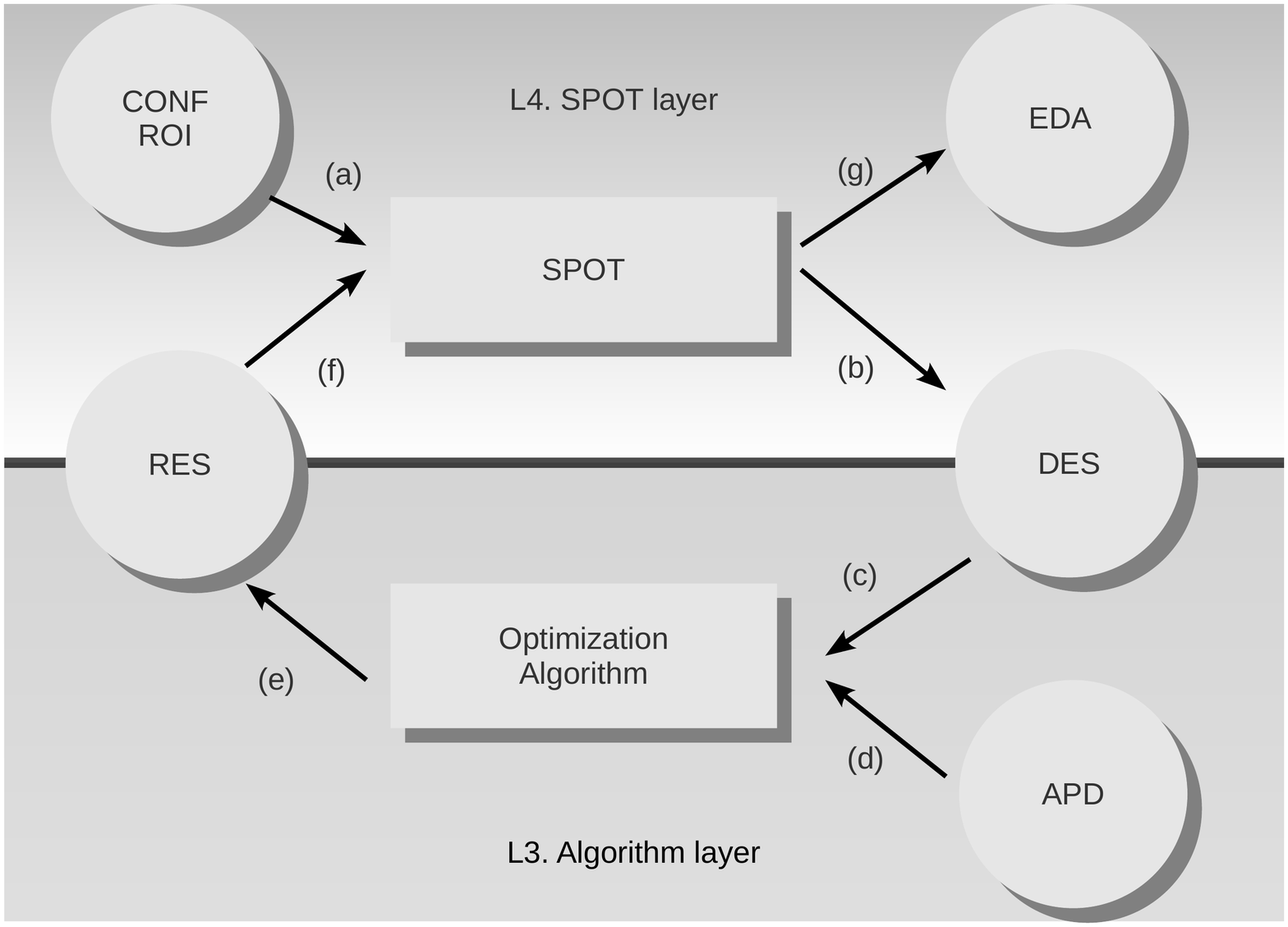}
\end{center}
\caption{SPOT interfaces.
The \SPOT loop can be described as follows:
\emph{Configuration}\/ (CONF) and \emph{region-of-interest}\/ (\ROI) files are
read by \SPOT (a). \SPOT generates a \emph{design}\/ (\DES) file (b).
The algorithm reads
the design file and
(c)
extra information, e.g., about the problem
dimension from the \emph{algorithm-problem design}\/ (\APD) file (d).
Output from the optimization algorithm are written to the \emph{result}\/
(\RES) file  (e). The result file is used by \SPOT to build the prediction model
(f). Data can be used by \emph{exploratory data analysis} (EDA)
tools to generate reports, statistics,
visualizations, etc. (g)}\label{fig:spotdataflow}
\end{figure}

\paragraph{Files Used During the Tuning Process} 
Each configuration file belongs to one \SPOT project, if the same basename is
used for corresponding files.
\SPOT uses
simple text files as interfaces from the algorithm to the statistical tools.
\begin{enumerate}
  \item The user has to provide the following files:
  \begin{enumerate}[(i)]
    \item   \emph{Region of interest} (\ROI)
    \index{region of interest}%
    files specify the region over which the algorithm parameters are tuned.
    Categorical variables such as the recombination operator in ES, can be 
    encoded as factors, e.g., ``intermediate recombination'' and ``discrete
    recombination.''
    \item \emph{Algorithm design} (\APD) files are used to specify parameters
    used by the algorithm, e.g., problem dimension, objective function, starting
    point, or initial seed.
    \item \emph{Configuration} files (\CONF) specify \SPOT specific parameters,
    such as the prediction model or the initial design size.
  \end{enumerate}
  \item \SPOT will generate the following files:
  \begin{enumerate}[(i)]
    \item \emph{Design\/} files (\DES) specify algorithm designs. They are
    generated automatically by \SPOT and will be read by the optimization algorithms.
    \item After the algorithm has been started with a parametrization from the
    algorithm design, the algorithm writes its results to the \emph{result
    file} (\RES). Result files provide the basis for many statistical
    evaluations/visualizations. They are read by \SPOT to generate prediction models. Additional prediction
    models can easily be integrated into \SPOT.
  \end{enumerate}
\end{enumerate}
Figure~\ref{fig:spotdataflow}
illustrates \SPOT interfaces and the data flow. 
The acronym EDA (exploratory data analysis) summarizes additional information
that can be used to add further statistical tools. For example, \SPOT writes a
best file (\BST), which summarizes information about the best configuration during the tuning process.
Note, that the problem design
can be modified, too. This can be done to analyze the robustness (effectivity)
of algorithms.

\begin{figure}
\begin{center}
\includegraphics[width=\linewidth]{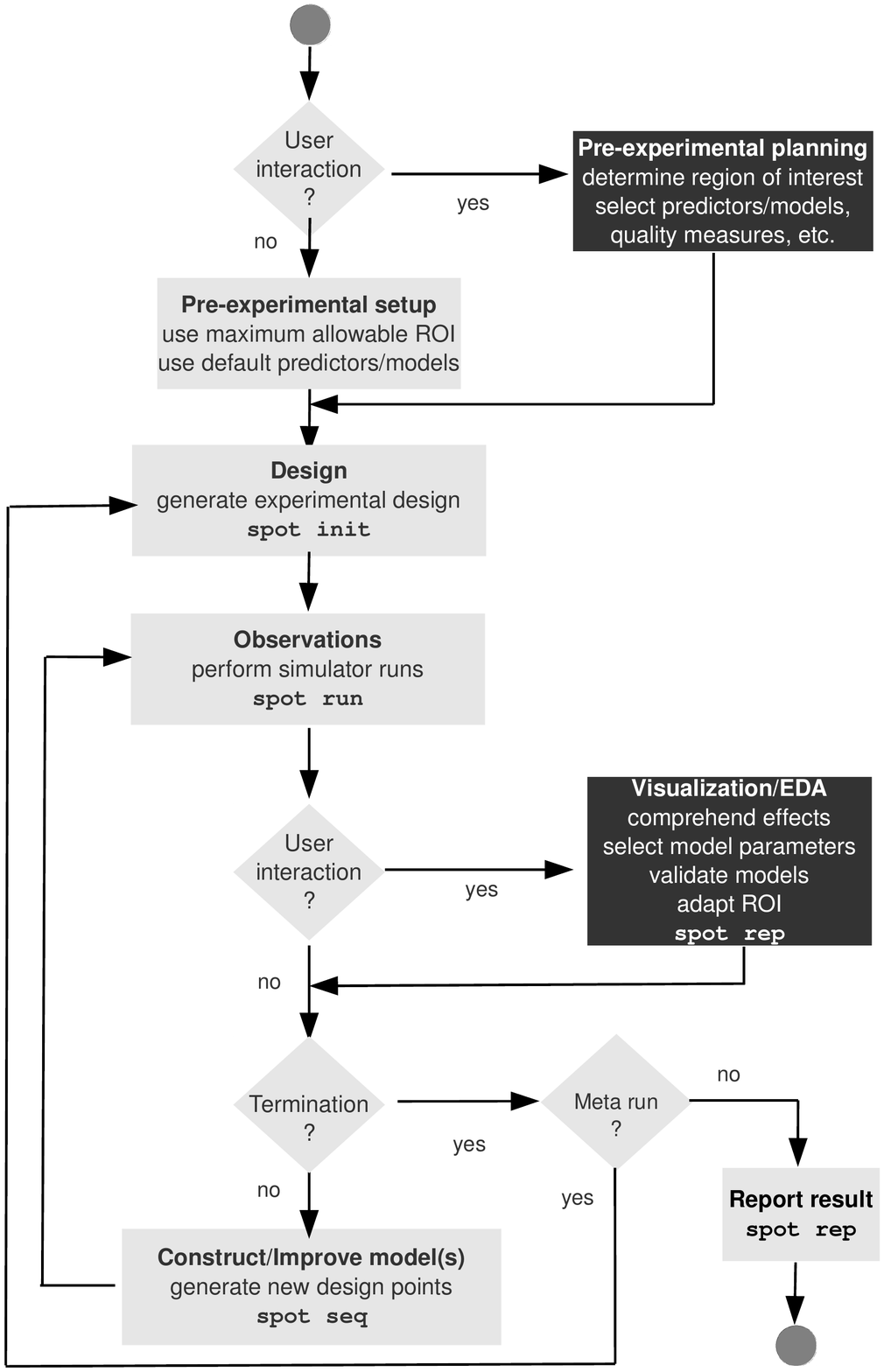}
\end{center}
\caption{The \SPOT process.
\emph{White font color}\/ indicates steps used in the interactive process only.
A \emph{typewriter font}\/ indicates the corresponding \SPOT commands.
To start the automatic mode, simply use the task {\tt auto}. Note that the
interaction points are optional, so \SPOT can be run without any user
interaction. \meta runs perform tuning processes for several configurations 
}\label{fig:spotProcess}
\end{figure}

\paragraph{SPOT Tasks}\label{sec:tasks}
\SPOT provides tools to perform
the following tasks (see also Fig.~\ref{fig:spotProcess}):
\begin{enumerate}
  \item \emph{Initialize}. An initial design is generated.  This is usually the
  first step during experimentation. The employed parameter region (\ROI) and
  the constant algorithm parameters (\APD) have to be provided by the user.
  \SPOT's parameters are specified in the \CONF file. Although it is
  recommended to use the same basename for \CONF, \ROI, and \APD files in order
  to define a project, this is not mandatory. \SPOT allows a flexible
  combination of different filenames, e.g., one \APD file can be used for
  different projects.
  \item \emph{Run.} This is
  usually the second step. The optimization algorithm is started with
  configurations of the generated design. Additionally information about the
  algorithms problem design are used in this step. The algorithm writes
  its results to the result file. 
  \item \emph{Sequential step}. A new
  design, based on information from the result file, is
  generated. A prediction model is used in this step. Several generic prediction models are available in \SPOT
  by default. To perform an  efficient analysis,  especially in
  situations when only few algorithms runs  are possible, user-specified prediction models can easily
  be integrated into \SPOT.
  \item \emph{Report}. An analysis, based on information from the result file,
  is generated. Since all data flow  is stored in files, new report facilities can
  be added  very easily. \SPOT contains some scripts to perform a basic
  regression analysis and plots such as histograms,  scatter plots, plots of the
  residuals, etc.
  \item \emph{Automatic} mode. In the automatic mode, the steps \emph{run}
  and \emph{sequential} are performed after an initialization for a
  predetermined number of times.
  \item \emph{Meta} mode. In the \meta mode, the tuning process is repeated for
  several configurations. For example, tuning can be performed for different
  starting points $\vecx_0$, several dimensions, or randomly chosen problem
  instances.
\end{enumerate}
As stated in Sect.~\ref{sec:practical}, \SPOT has been applied to several
optimization tasks which might give further hints how \SPOT can be used.
Bartz-Beielstein~\cite{Bart09e,Bart09f} present case studies that may serve
as good starting points for \SPOT applications.

\section{Details}\label{sec:details}
We will discuss functions which are used during the four \SPOT steps initialize,
run, sequential, and report.

\subsection{Initialize}
During this step, the initial design is generated and written to the design
file.
{\tt spotCreateDesignLhs}, which is based on \R's {\tt lhs} package, is
recommended as a simple space filling design.

Alternatively, factorial designs can be used.
{\tt spotCreateDesignFrF2}, which is based on Groemping's  {\tt FrF2} package,
see \url{http://cran.r-project.org/web/packages/FrF2/index.html},
generates a fractional factorial design with center point.

Furthermore, the number of initial design
points, the type of the experimental design etc. have to be specified before
the first \SPOT run is performed. These information are stored in the
\emph{configuration file}\/(\CONF), see Listing~\ref{lst:demo71}.
\lstset{caption ={demo07RandomForestSann.conf}\label{lst:demo71}}
\begin{lstlisting}[firstline=21]
alg.func = "spotAlgStartSann"
auto.loop.nevals = 100
init.design.func = "spotCreateDesignLhs"
init.design.size = 10
init.design.repeats  = 2
seq.predictionModel.func = "spotPredictRandomForest"
\end{lstlisting}
The configuration file plays a central role in \SPOT's tuning process. It
stores information about the optimization algorithm ({\tt alg.func}) and the
meta model ({\tt seq.predictionModel.func}). \SPOT uses a
classification scheme for its variables:
{\tt init} refers to variables which were used during the initialization step,
{\tt seq} are variables used during the sequential step, and so forth.

The experimental region is specified in the \emph{region of interest}\/ (\ROI)
file, see Listing~\ref{lst:roi71}.
In the {\tt demo07RandomForestSann} project, two numerical variables with
values from the interval $[1;50]$ are used. \SPOT employs a mechanism which adapts the
region of interest automatically. Information about the actual region of
interest are stored in the {\tt aroi} file, which is handled by \SPOT
internally.
\lstset{caption ={demo07RandomForestSann.roi}\label{lst:roi71}}
\begin{lstlisting}[firstline=1]
name low high type
TEMP 1 50 FLOAT
TMAX 1 50 FLOAT
\end{lstlisting}
Now all the source files are available. In order to generate the
initial design, simply call \SPOT as follows.
\begin{Schunk}
\begin{Sinput}
spot("demo07RandomForestSann.conf","init")
\end{Sinput}
\end{Schunk}
The ``init'' call generates a \emph{design file}\/ (\DES), which is shown in
Listing~\ref{lst:des71}.
\lstset{tabsize=2,caption ={Design file demo07RandomForestSann.des generated by
\SPOT}\label{lst:des71}}
\begin{lstlisting}[firstline=1]
TEMP                TMAX			CONFIG REPEATS STEP SEED
35.6081542731496	20.5193298289552	 1	2		0	1235
3.03074611076154	30.9299647800624	 2	2		0	1235
35.0430960887112	11.6981793923769	 3	2		0	1235
18.7132237656275	49.8082008753903	 4	2		0	1235
13.9964890434407	35.3148515065433	 5	2		0	1235
26.2654501354089	25.7817166472087	 6	2		0	1235
24.1049260527361	3.67511987574399	 7	2		0	1235
7.34134425916709	25.289775890857		 8	2		0	1235
49.035177000449		42.5625789203448	 9	2		0	1235
42.5434358732775	7.34647449669428	10	2		0	1235
\end{lstlisting}
Since we have chosen the SPOT plugin {\tt spotCreateDesignLhs},
a Latin hypercube design is generated.
Each configuration is labeled with a configuration number. The column REPEATS
contains information from the variable {\tt init.design.repeats}. Since no meta
model has been created yet, STEP is set to $0$  for each configuration.
Finally, the SEED, which is used by the algorithm, is shown in the last column.

\subsection{Run}
Parameters from the design file are read and the algorithm is executed.
Each run results in one fitness value (single-objective optimization) or
several values (multi-objective optimization). Fitness values with
corresponding parameter settings are written to the result file.
The user has to set up her own interface for her algorithm $A$. Examples are
provided, see Sect.~\ref{sec:run}.
The command 
\begin{Schunk}
\begin{Sinput}
spot("demo07RandomForestSann.conf","run")
\end{Sinput}
\end{Schunk}
executes the \run task.
Results from this run are written to the \emph{result file}\/(\RES), which is
shown in Listing~\ref{lst:res71}.
\lstset{caption ={demo07RandomForestSann.res}\label{lst:res71}}
\begin{lstlisting}[firstline=1]
Y TEMP TMAX FUNCTION DIM SEED CONFIG STEP
3.30597157332377 35.6081542731496 21 Branin0.0 2 1235 1 0
5.0386282545543 35.6081542731496 21 Branin0.0 2 1236 1 0
0.400306954041771 3.03074611076154 31 Branin0.0 2 1235 2 0
0.398257998573415 3.03074611076154 31 Branin0.0 2 1236 2 0
0.445520325383134 35.0430960887112 12 Branin0.0 2 1235 3 0
2.226649807247 35.0430960887112 12 Branin0.0 2 1236 3 0
0.497684708317145 18.7132237656275 50 Branin0.0 2 1235 4 0
1.83495383747858 18.7132237656275 50 Branin0.0 2 1236 4 0
0.399119231011623 13.9964890434407 35 Branin0.0 2 1235 5 0
0.542312919825205 13.9964890434407 35 Branin0.0 2 1236 5 0
2.72322377892531 26.2654501354089 26 Branin0.0 2 1235 6 0
0.417190547410852 26.2654501354089 26 Branin0.0 2 1236 6 0
4.7533857289203 24.1049260527361 4 Branin0.0 2 1235 7 0
1.61981739130332 24.1049260527361 4 Branin0.0 2 1236 7 0
0.403447629608046 7.34134425916709 25 Branin0.0 2 1235 8 0
0.411160853072728 7.34134425916709 25 Branin0.0 2 1236 8 0
2.98491602241286 49.035177000449 43 Branin0.0 2 1235 9 0
5.24485760127901 49.035177000449 43 Branin0.0 2 1236 9 0
6.91947230812718 42.5434358732775 7 Branin0.0 2 1235 10 0
0.521764744656569 42.5434358732775 7 Branin0.0 2 1236 10 0
\end{lstlisting}

\subsection{Sequential}
Now that results have been written to the result file, the meta model can be
build.
 \begin{Schunk}
\begin{Sinput}
spot("demo07RandomForestSann.conf","seq")
\end{Sinput}
\end{Schunk}
The sequential call generates a new \emph{design file}\/ (\DES), which is shown
in Listing~\ref{lst:des72}.
\lstset{caption ={Design file demo07RandomForestSann.des generated by
\SPOT}\label{lst:des72}}
\begin{lstlisting}[firstline=1]
TEMP TMAX CONFIG REPEATS repeatsLastConfig STEP SEED
3.03074611076154 31 2 1 2 1 1237
6.13796767716994 31.5014664068934 11 3 3 1 1235
1.50800024462165 28.6290849421476 12 3 3 1 1235
\end{lstlisting}
In order to improve confidence, the best solution found so far is evaluated
again. To enable fair comparisons, new configurations are evaluated as many
times as the best configuration found so far.
Note, other update schemes are possible.

If \SPOT's budget is not exhausted, the new configurations are evaluated, i.e.,
\run is called again, which updates the result file. In the following step,
\seq is called again etc.

To support exploratory data analysis, \SPOT also generates a best file, 
which is shown
in Listing~\ref{lst:bst72}.
\lstset{caption ={Best file demo07RandomForestSann.bst generated by
\SPOT}\label{lst:bst72}}
\begin{lstlisting}[firstline=1]
Y TEMP TMAX COUNT CONFIG
0.398592091098704 3.03074611076154 31 2 2
0.39950017406572 1.8086370804091 31 3 11
0.399448475332373 1.8086370804091 31 4 11
0.399732585200016 1.8086370804091 31 5 11
0.399443742300062 1.8086370804091 31 6 11
0.400057920171103 3.03074611076154 31 3 2
0.400239513036257 1.8086370804091 31 7 11
0.400845187050665 1.8086370804091 31 9 11
0.400892337114249 3.82960996863898 30 4 13
0.401138794573396 1.8086370804091 31 10 11
0.399842856102580 1.20055107064079 31 10 29
0.399842856102580 1.20055107064079 31 10 29
0.399842856102580 1.20055107064079 31 10 29
\end{lstlisting}
The best file is updated after each \SPOT iteration and be be used for an
on-line visualization tool, e.g., to illustrate search progress or stagnation,
see Fig.~\ref{fig:final}. The variable COUNT reports the number of REPEATS used
for this specific configuration.
 
\subsection{Report}
If \SPOT's termination criterion is fulfilled, a report is
generated. By default, \SPOT provides as simple report function which reads data
from the res file and produces the following output:
\begin{Schunk}
\begin{Soutput}
Best solution found with 223 evaluations:
     Y     TEMP TMAX COUNT CONFIG
0.3998429 1.200551   31    10     29
\end{Soutput}
\end{Schunk}

\subsection{Automatic}
\SPOT's \auto task performs steps \init, \run, \seq, \run,
\seq, etc. until the termination criterion is fulfilled, see
Fig.~\ref{fig:spotProcess}.
It can be invoked from \R's command line via
\begin{Schunk}
\begin{Sinput}
spot("demo07RandomForestSann.conf","auto")
\end{Sinput}
\end{Schunk}

\section{Plugins}\label{sec:plugins}
\SPOT comes with a basic set of \R functions for generating initial designs,
starting optimization algorithms, building meta models, and generating reports.
This set can easily be extended with user defined \R functions, so called
plugins. Further plugins will be added in forthcoming \SPOT versions. Here, we describe
the interfaces that are necessary for integrating user-defined plugins into
\SPOT.
\subsection{Initialization Plugins}
The default plugin for generating an initial design is 
{\tt init.design.func = "spotCreateDesignLhs"}.
It uses information about the size of the initial design {\tt
init.design.size}. The number $n$ of design variables $x_i$ ($i=1,\ldots,n$),
their names {\tt pNames}, and their ranges $a_i \leq x_i\leq b_i$ can be
determined with \SPOT's internal {\tt alg.roi} variable, which is
passed to the initialization plugin.
\begin{Schunk}
\begin{Sinput}
> pNames <- row.names(alg.roi);
> a <-  alg.roi[ ,"low"];
> b <-  alg.roi[ ,"high"];
\end{Sinput}
\end{Schunk}	
Note, {\tt pNames}, $a$, and $b$ are vectors of size $n$. 
Based on this information, a data frame with initial design points is generated.
For example, the data frame from the {\tt demo07RandomForestSann}  project
reads:
\begin{Schunk}
\begin{Soutput}
        TEMP      TMAX
1  35.608154 20.519330
2   3.030746 30.929965
3  35.043096 11.698179
4  18.713224 49.808201
5  13.996489 35.314852
6  26.265450 25.781717
7  24.104926  3.675120
8   7.341344 25.289776
9  49.035177 42.562579
10 42.543436  7.346474
\end{Soutput}
\end{Schunk}
These values are written to the initial design file, see
Listing~\ref{lst:des71}. The reader is referred to the {\tt
spotCreateDesingLhs} function for further details.

The plugin {\tt spotCreateDesignFrF2} generates a central composite design
and can be used as a template for fractional factorial design plugins.
Currently, \SPOT implements the following \init plugins:
\begin{itemize}
  \item {\tt spotCreateBasicDoe3R}: creates a fractional-factorial
  design (resolution III)
  \item {\tt spotCreateFrF2}:  creates a resolution III design with
  center point and star points
  \item {\tt spotCreateLhs}:  creates a Latin hypercube design
\end{itemize}
Note, these plugins should be used as templates and can be easily adopted to
specific situations.

\subsection{Run Plugins}\label{sec:run}
The \run plugin {\tt spotAlgStartSann}, which is used
as an interface to \R's \SANN algorithm, is shown in the Appendix,  see Listing~\ref{lst:run1}.
Basically, the user has to specify variable names to be read from the design
file, see Sect.~\ref{sec:desread}, and written to the result file, see
Sect.~\ref{sec:reswrite}, and the call of the algorithm, see
Sect.~\ref{sec:exec}.

\subsubsection{Reading Values From the Design File}\label{sec:desread}
To add a new variable, say COLOR, the user simply adds the following line of
code to the run file:
\begin{Schunk}
\begin{Sinput}
if (is.element("COLOR", pNames)){color <- des$COLOR[k]}
\end{Sinput}
\end{Schunk}

\subsubsection{Executing the Algorithms}\label{sec:exec}
Next, the call of the algorithm has to be specified. In our
example, 
\begin{Schunk}
\begin{Sinput}
y <- optim(x0, spotFunctionBranin, method="SANN",
control=list(maxit=maxit, temp=temp, tmax=tmax, parscale=parscale, 
color=color))
\end{Sinput}
\end{Schunk}

\subsubsection{Writing Results to the Result File}\label{sec:reswrite}
And finally, in order to write the variable to the result file, it has to be
added to the following list:
\begin{Schunk}
\begin{Sinput}
res <- list(Y = y, TEMP = temp, TMAX = tmax, 
COLOR = color, SEED = seed, CONFIG = conf)
\end{Sinput}
\end{Schunk}

\subsubsection{Interfacing With Algorithms Written in Other Programming
Languages} 

We will demonstrate how JAVA programs can be called from \SPOT.
The procedure consists of two steps: First, a call string is build. Then, \R's
{\tt system} function is used for executing the callString.
\begin{Schunk}
\begin{Sinput}
callString <- paste("java -jar simpleOnePlusOneES.jar", 
seed, steps, target, f, n, xp0, sigma0, a, g, px, py, sep = " ")		
y <-system(callString, intern= TRUE)
\end{Sinput}
\end{Schunk}
This procedure can be applied to any optimization algorithm.
Templates for state-of-the-art optimization algorithms will be added to
forthcoming \SPOT version. Users are encouraged to submit interfaces to their
algorithms to the \SPOT development team.

Currently, \SPOT implements the following \run plugins:
\begin{itemize}
  \item {\tt spotAlgStartSann:} Interface to \R's simulated annealing \SANN
  \item {\tt spotAlgStartES:} Interface to an \ES based on~\citet{BeSw02a}
  \item {\tt spotFuncStartBranin:} Interface to the Branin function. \SPOT is
  used as an optimizer, not as a tuner, see also Sect.~\ref{sec:opt}
\end{itemize}

Additional \run packages are available, e.g., 
\begin{Schunk}
\begin{Sinput}
> demo(spotDemo11Java, ask=FALSE) 
\end{Sinput}
\end{Schunk}
demonstrates how a (1+1)-\ES, which is implemented in Java, can be
tuned with \SPOT.

\subsection{Sequential Plugins}\label{sec:seq}
During \SPOT's sequential step one or several meta models are generated. These
models use information from the result file. New, promising design points are
generated. Therefore, a large number of randomly generated design points are
evaluated on the meta model. Configurations with the best estimated
objective function values  are written
to the design file and will be evaluated during the \run step, see line~\ref{spo-alg:d} in Algorithm~\ref{alg:spo-algorithm}.

\SPOT provides two types of data assembled from the result file:
\emph{Raw}\/ data comprehend parameter values $\vecx$ (configurations) and
related objective function values $y$, whereas \emph{merged}\/ data map same
$\vecx_i$ configurations to one configuration $\vecx_j$. The corresponding $y_i$
values are merged according to the merge function (default: mean), e.g., $y_j=
\sum_1^n y_i/n$. In our example, the random forest is generated with raw data.
\R's generic {\tt predict} function is used to evaluate new data on the meta model (random forest). Finally, the best design points are determined.

The random forest meta model is implemented as shown in
Listing~\ref{lst:rf1}.
\lstset{caption={spotPredictRandomForest.R}\label{lst:rf1}}
\begin{lstlisting}[firstline=17]
spotPredictRandomForest <- function(rawB,mergedB,largeDesign,spotConfig){	
spotInstAndLoadPackages("randomForest")	
xNames <- setdiff(names(rawB),"y")
x <- rawB[,xNames]; y <- rawB$y	
fit <- randomForest(x, y)		
res <- predict(fit,largeDesign)
largeDesign <-  largeDesign[order(res,decreasing=FALSE),]	
newDesign <- largeDesign[1:spotConfig$seq.design.new.size,]}
\end{lstlisting}
Currrently (June 2010), \SPOT provides interfaces to the following meta
modeling approaches:
\begin{itemize}
  \item Regression models ({\tt lm}; {\tt rsm}): 
  \begin{enumerate}
    \item {\tt spotPredictLm}
    \item {\tt spotPredictLmOptim}
  \end{enumerate}
    \item Tree based models ({\tt tree}; {\tt randomForest})
\begin{enumerate}
    \item {\tt spotPredictTree}
    \item {\tt spotPredictRandomForest}
  \end{enumerate}
  \item Gaussian process models ({\tt mlegp}; {\tt tgp})
  \begin{enumerate}
    \item {\tt spotPredictTgp}
    \item {\tt spotPredictMlegp}
  \end{enumerate}      
\end{itemize}
The Appendix presents an example
(Listing~\ref{lst:rfmlegp1}) how several meta models can be combined.
Interfaces to further meta models will be provided in future releases of the
\SPOT package.

\subsection{Report Plugins}

\begin{figure}
\includegraphics[width=\linewidth]{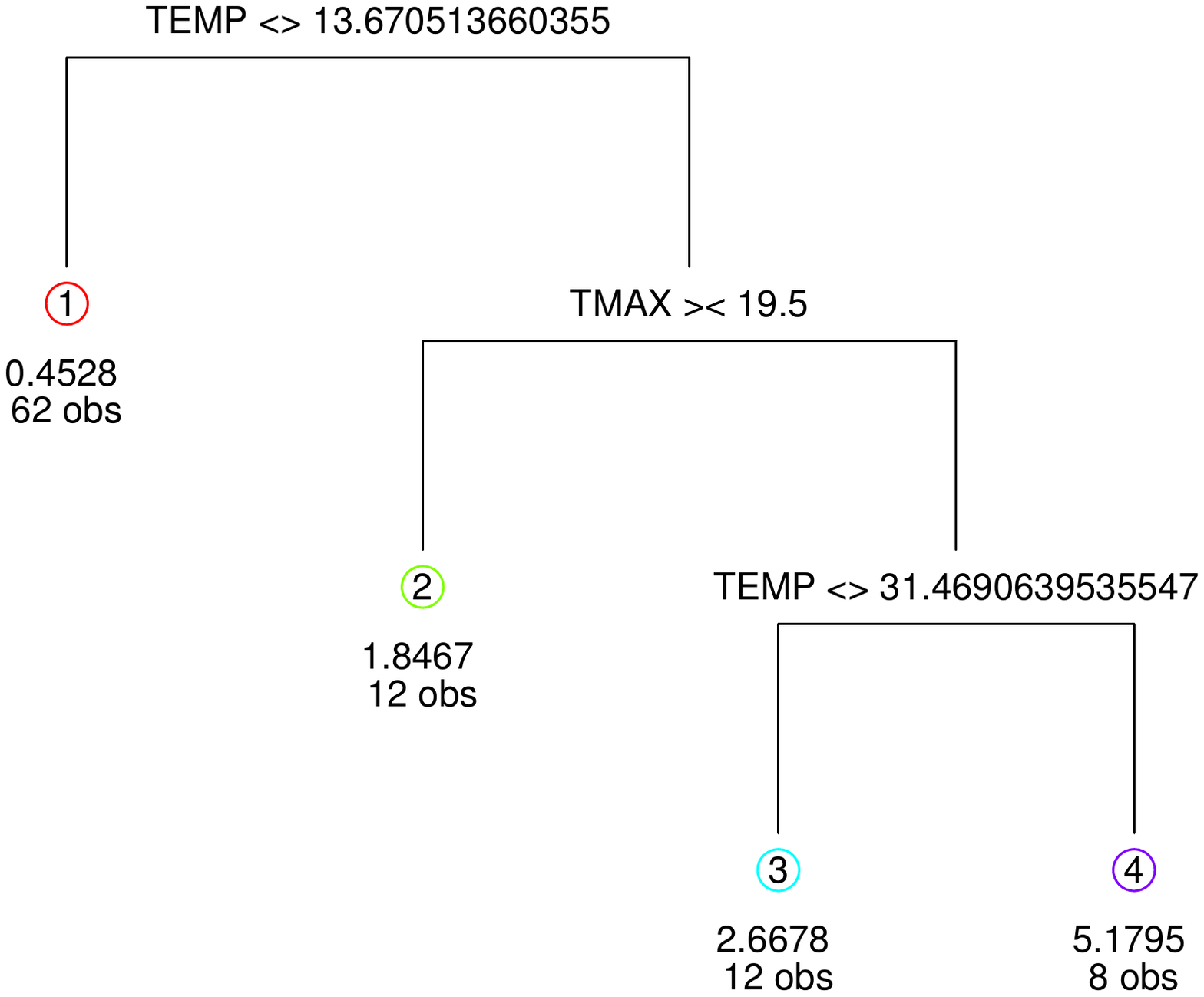}
\caption{Tuning \SANN with \SPOT. An \rsm and \tree based approach are combined.
Similar to the random forest based meta modeling, TEMP has the largest effect}
\label{fig:tune3}
\end{figure}

\SPOT comes with a simple report plugin {\tt spotReportDefault.R}.
It reports the best configuration from the tuning procedure, illustrates the
tuning process (evolution of the best solution as shown in
Figs.~\ref{fig:tune1} and \ref{fig:tune2}), and generates a simple regression
tree as shown in Fig.~\ref{fig:tune3}.

\begin{figure}
\centering
\includegraphics[width=0.75\linewidth]{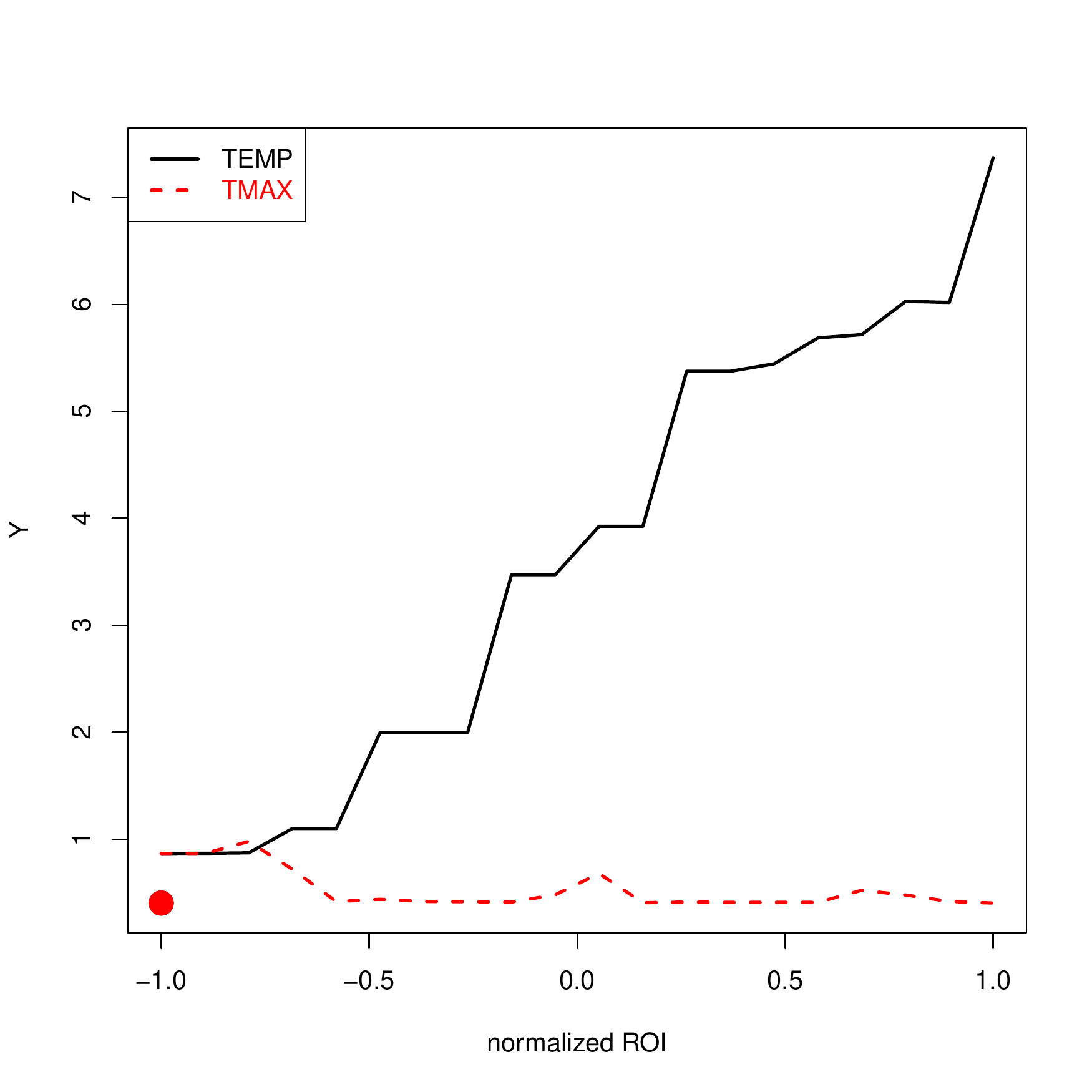}
\caption{Results from the \SANN tuning procedure. Function values
$Y$ plotted versus parameter values. \randomForest was used to predict values
for one variable, say \temp, while the other variable (\tmax) was set its
optimal value. Values for both variables were normalized. This plot was
generated with the {\tt spotReportSens} plugin
\label{fig:sens1} }
\end{figure}

\begin{figure}
\centering
\includegraphics[width=\linewidth]{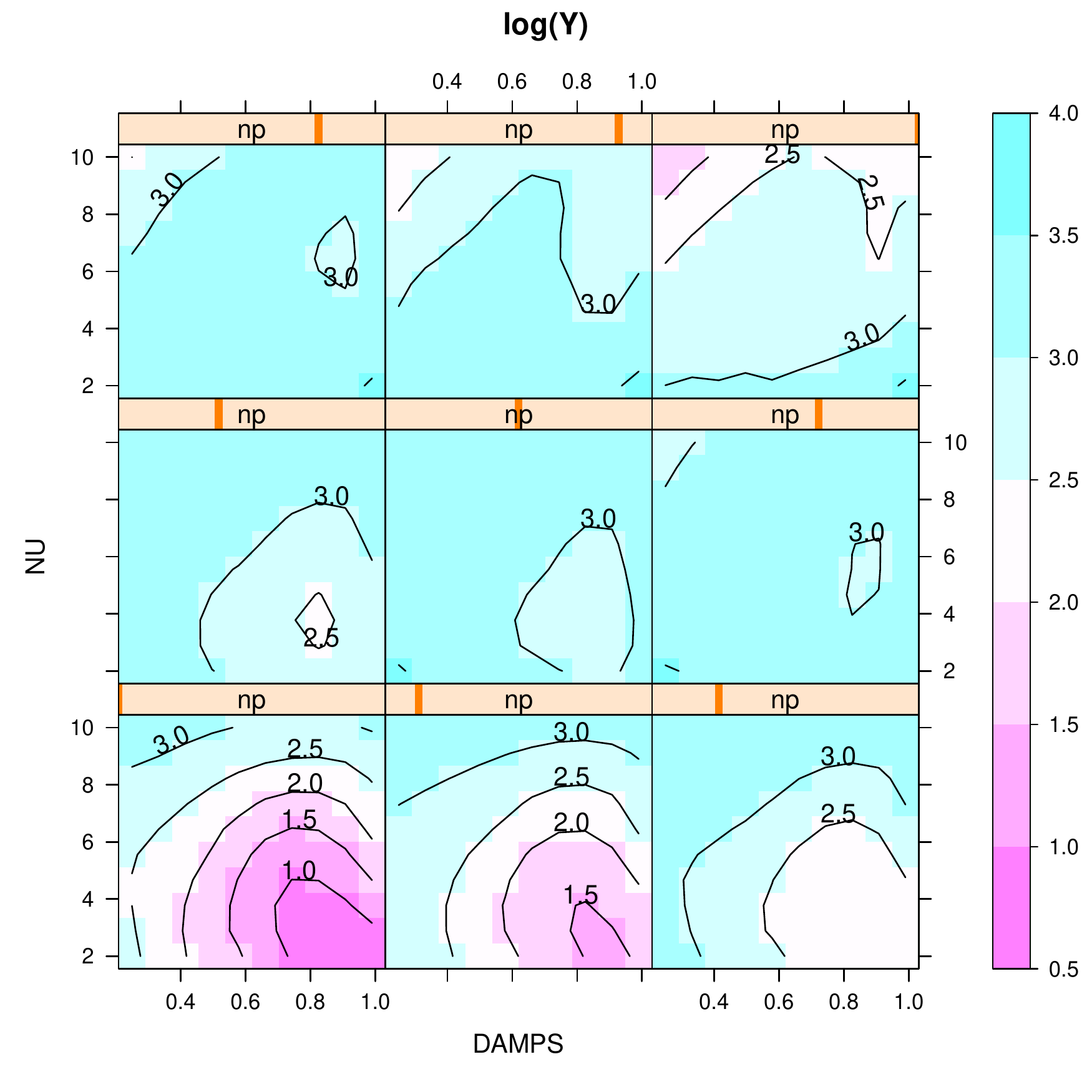}
\caption{This figure shows an \EDA example taken from~\cite{Bart09f}. Contour
plots based on 82 function evaluations of the CMA evolution strategy (CMA-ES)
optimizing the Ackley function are shown~\citep{Hans06a}. Smaller values are better. Better
configurations are placed in the lower area of the panels. The CMA-ES has
four algorithm parameters (CS, NU, DAMPS, and NPARENTS). The parameter CS
is held constant. NU is plotted versus DAMPS, while values of the
parameter NPARENTS (np), are varied with the slider on top of each panel
\label{fig:cnt0015Ackley}}
\end{figure}

User defined report functions can easily be added.
Wolfgang Konen has written a report plugin which uses \randomForest to
visualize factor effects, see Fig.~\ref{fig:sens1}.
Figure~\ref{fig:cnt0015Ackley} demonstrates how \EDA tools can be applied to
analyse effects and interactions.
Note, results from the
result file can be used for detailed reports.
At this stage, EDA tools are the method of choice.

\section{Refinement}\label{sec:refine}
\subsection{Combining Meta Models And Adaptation of the Region of Interest}
During the sequential step, \SPOT can use different meta models. 
The following example demonstrates how results from tree based regression and
response surface modeling can be combined.

\subsubsection{Designs}
A \emph{central composite design}\/ (CCD) was chosen as the starting point of
the tuning process.
\SPOT's {\tt spotCreateDesignFrF2} plugin can be used to generate design points.
After the first run is finished, we can use \SPOT's report facility to analyze
results. Since we have chosen a classical factorial design, we will use
\emph{response-surface methodology}\/ (\RSM).
\citet{Lent09a} describes an implementation of \RSM in {\tt R}.
This \R package \rsm has many
useful tools for an analysis of the results from the \SPOT runs.
After evaluating the algorithm in these design points, a seccond order 
regression model with interactions is fitted to the data. Functions from the
\rsm package were used by the \SPOT plugin {\tt spotPredictLmOptim}.
Before meta models are build, data are standardized. Data in the original units
are mapped to coded data, i.e., data with values in the interval $[-1,1]$.

\subsubsection{Response Surface Models}
Based on the number of design points, \SPOT automatically determines whether a
first-order, two-way interaction, pure quadratic, or second order model can
 be fitted to the data.
The CCD generated by {\tt spotCreateDesignFrF2} allows the fit of an
second-order model which can be summarized as follows.
\begin{Schunk}
\begin{Soutput}
Coefficients:
            Estimate Std. Error t value Pr(>|t|)   
(Intercept)  0.93070    0.46212   2.014   0.1375   
x1           2.29074    0.36382   6.296   0.0081 **
x2          -1.98286    0.36307  -5.461   0.0121 * 
x1:x2       -2.29498    0.40674  -5.642   0.0110 * 
x1^2         1.86950    0.87244   2.143   0.1215   
x2^2        -0.05832    0.87337  -0.067   0.9510   
---
Residual standard error: 0.8135 on 3 degrees of freedom
Multiple R-squared: 0.9734,	Adjusted R-squared: 0.9291 
F-statistic: 21.97 on 5 and 3 DF,  p-value: 0.01437 
\end{Soutput}
\end{Schunk}

\subsubsection{Using Gradient Information}\label{sec:grad}
\begin{figure}
\centering
\includegraphics[width=0.75\linewidth]{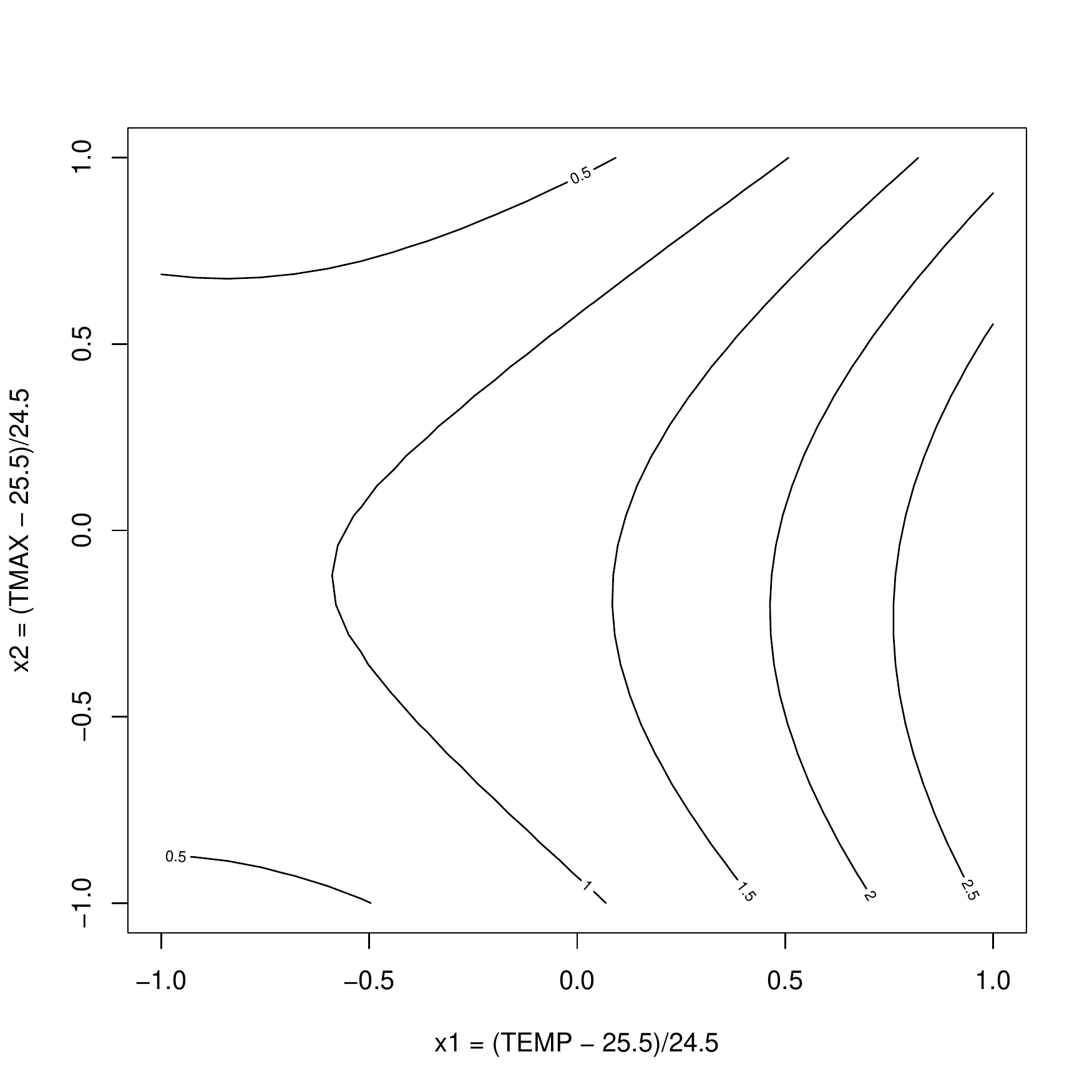}
\caption{Response surface model based on the initial design\label{fig:rsm1}.
\rsm was used to generate this plot}
\end{figure}

The response surface analysis determines the following stationary point on
response surface: $(-0.8447783, -0.3781626)$, or, in the original
 units \temp = 4.802932 and \tmax = 16.235016. The eigenanalysis shows
that 
the eigenvalues ($\lambda_1 =2.4042085$; $\lambda_2 =-0.5930265$) have different
signs, so this is a saddle point, as can also be seen in Fig.~\ref{fig:rsm1}.
\SPOT automatically determines the path of the steepest descent and selects
five points, using the old center point as a starting point, in its direction:
(23.7605, 27.215),
(22.315, 29.224), 
(21.4085, 31.6005), 
(21.139, 34.271), and 
(21.4085, 37.0395), i.e., decreasing \temp and increasing \tmax values are
chosen.
Rather than at
the origin, \SPOT can start the search at the saddle point. 
Set {\tt seq.useCanonicalPath = TRUE} to enable this feature. 
In this case, \SPOT
determines the most steeply rising ridge in both directions, see also
\citet{Lent09a} for details:
\begin{Schunk}
\begin{Soutput}
dist     x1     x2 |   TEMP    TMAX 
1 -0.2 -0.760 -0.197 | 6.8800 20.6735
2 -0.1 -0.803 -0.288 | 5.8265 18.4440
3  0.0 -0.845 -0.378 | 4.7975 16.2390
4  0.1 -0.887 -0.469 | 3.7685 14.0095
5  0.2 -0.929 -0.559 | 2.7395 11.8045
\end{Soutput}
\end{Schunk}
In addition to the points from the steepest descent, the best point from the
first design (1,50) is evaluated again. 
Now, these points are evaluated and a new \rsm model is build.

\subsubsection{Automatic Adaptation of the Region of Interest}
\SPOT modifies the region of interest, if {\tt seq.useAdaptiveRoi = TRUE}.
This procedure consists of two phases, which are repeated in an
alternating manner.

During the \emph{orientation} phase, the direction of the largest improvement
is determined as described in Sect.~\ref{sec:grad}. Based on an existing design
and related function values, the path of the steepest descent is determined. A small number of points is chosen
from this path. Optimization runs are performed on these design points.
In some situations, where no gradient information is available, the best point
from a large number of design points, which were evaluated on the regression
model, is chosen as the set of improvement points.

The \emph{recalibration} phase determines the best point $\vecx_b$. It can be
selected from the complete set of evaluated design points or from the points along the
steepest descent only.
The best point $\vecx_b$ defines the new center point of a central composite
design. The minimal distance of $\vecx_b$ to the borders of the actual region
of interest defines the radius of this design.
If $\vecx_b$ is located at (or very close to) the borders of the region of
interest, a Latin hypercube design which covers the whole region of interest is
determined. This can be interpreted as a restart.
To prevent premature convergence of this procedure, one additional new
design point is generated by a tree based model.

\begin{figure}
\includegraphics[width=\linewidth]{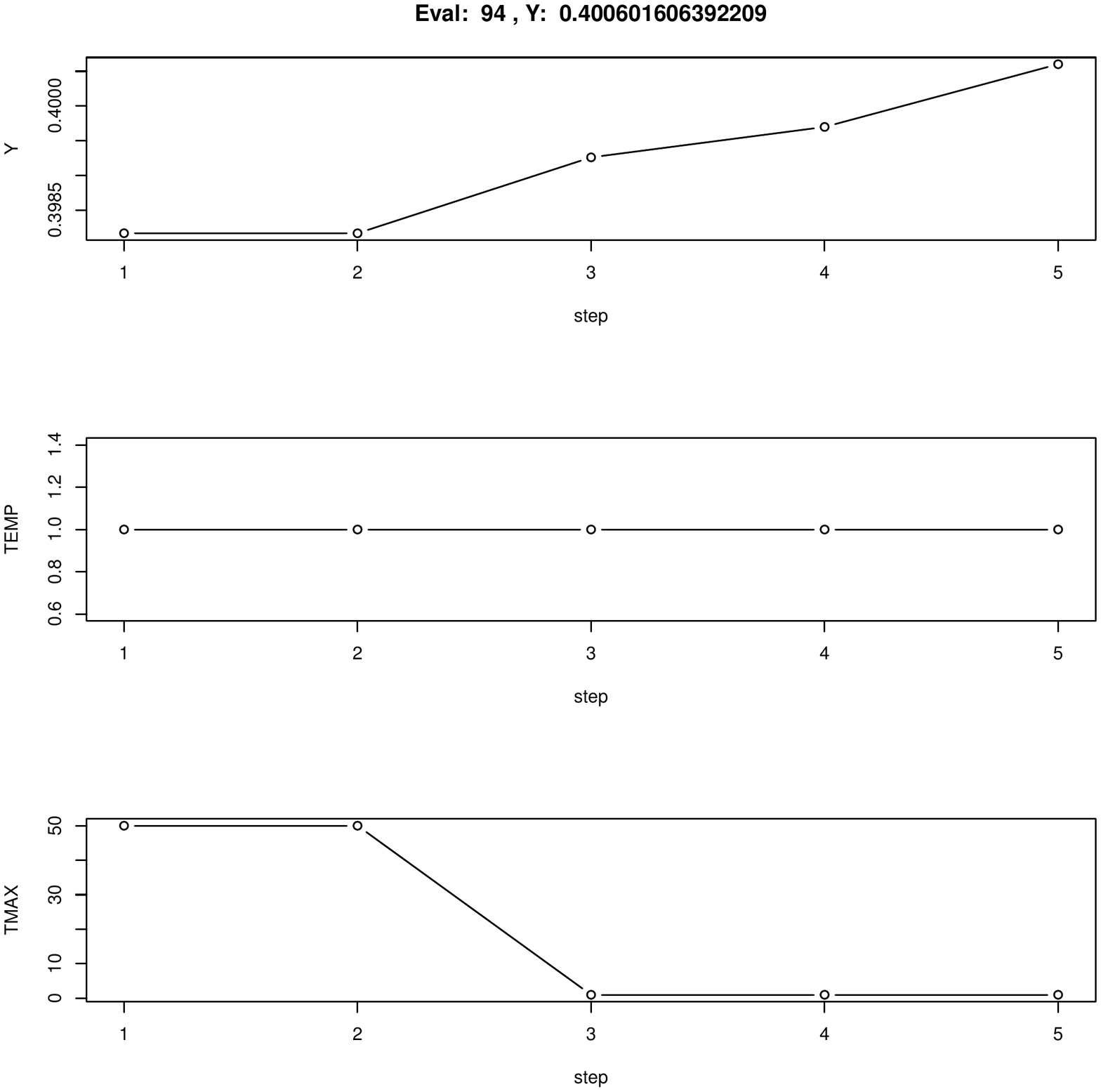}
\caption{Tuning \SANN with \SPOT. An rsm and tree based approach are combined}
\label{fig:tune2}
\end{figure}

\begin{figure}
\begin{center}
\includegraphics[width=0.5\linewidth]{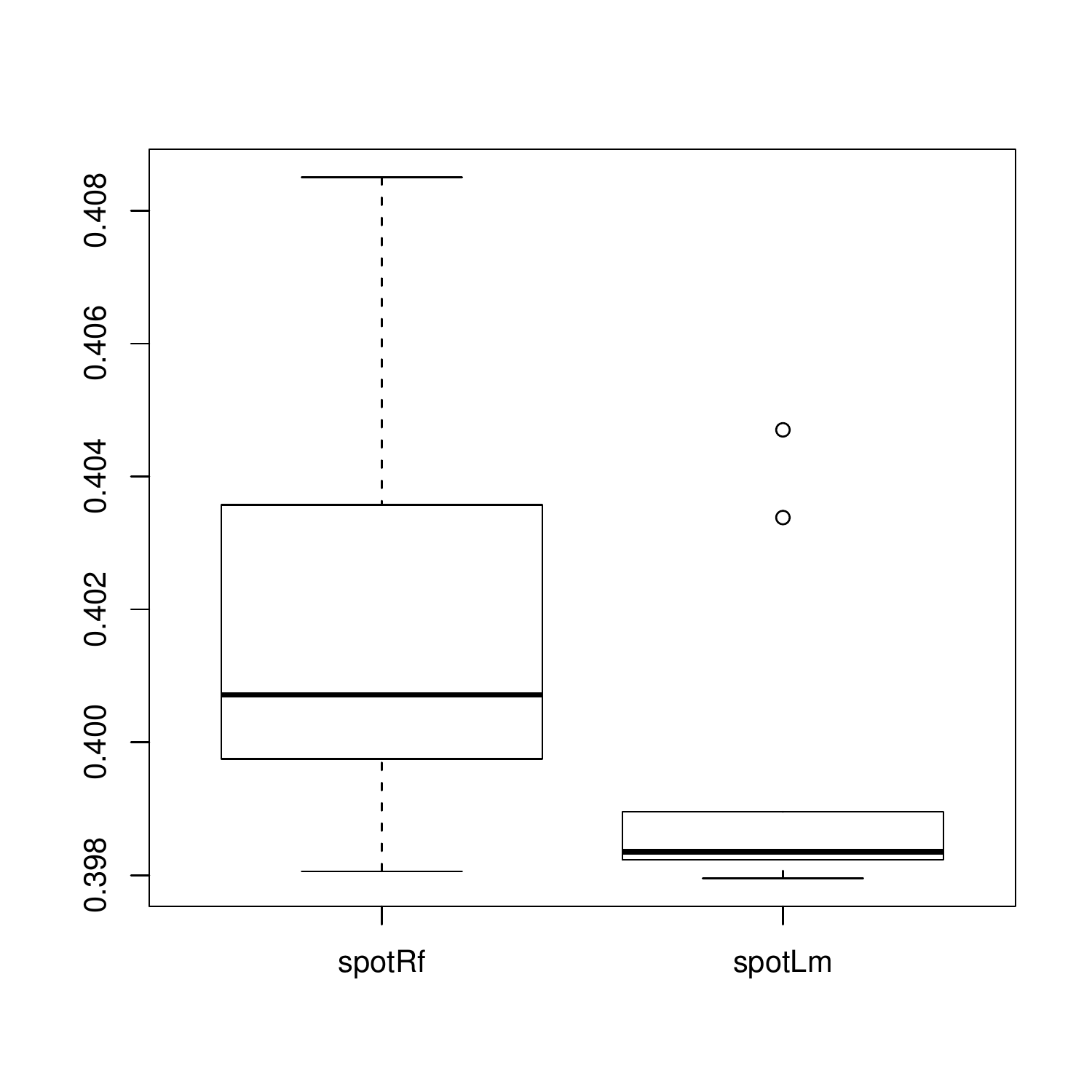}
\end{center}
\caption{Comparison of \SANN's random forest tuned parameter values with
parameter settings obtained with \rsm  (\emph{spotLm})
\label{fig:box2}}
\end{figure}

Next, the orientation phase is repeated.
The tuning process with adaptive ROI is visualized in Fig.~\ref{fig:tune2}.
The final output from this tuning process, which is based on regression models
and tree based regression reads:
\begin{Schunk}
\begin{Soutput}
Best solution found with 94 evaluations:
          Y TEMP TMAX COUNT CONFIG
0.4006016    1    1     6      2
\end{Soutput}
\end{Schunk}
As in Sect.~\ref{sec:tunespot}, ten repeats of the best solution from this
tuning process are generated. Results are shown in Fig.~\ref{fig:box2}.
Note, this result was found with only half of the number of \SANN runs compared
to the random forest modeling approach from Sect.~\ref{sec:tunespot}.
The example demonstrates how the usage of gradient information can
accelerate the tuning procedure.

\subsection{Numerical and Categorical Values}\label{sec:cate}
\SPOT provides mechanisms for handling type information.
Categorical values such as ``red'', ``green'', and ``blue'' have to be coded as
integer values, e.g., ``1'', ``2'', and ``3''  in the \ROI file. By default,
they are treated as numerical values (\FLOAT). 
They can be treated as
\emph{factors}, if the corresponding type information
(\FACTOR) is provided in the type column of the \ROI file.
Alternatively, the type \INT can be specified in the \ROI file. These parameters
are treated as numerical values, but the {\tt spotCreateDesign} plugins
generate integer values which are written to the design files.
\citet{Bart09f} presents an example which illustrates the usage of type
information in \SPOT.

\subsection{Meta Projects}
\SPOT allows the definition of \emph{meta projects}. Meta project perform tuning
over a set of problem instances. One interesting task is to analyze
interactions between the search-space dimension, say $d$, and the best algorithm
design $\vecp^*$. For example, the experimenter can search for dependencies
between population size in \ES and $d$. For a detailed documentation the reader
is referred to the package help manuals.

\subsection{\SPOT as an Optimization Algorithm}\label{sec:opt}
\SPOT itself can be used as an optimization algorithm. The package includes some
demos to illustrate this feature. For example, {\tt spotDemoLm03Branin} uses a
linear (meta) model to optimize Branin's function.

\section{Summary and Outlook}\label{sec:summary}
This article present basic features of the \SPOT package which is implemented
in \R.
\SPOT provides tools for automatic and interactive tuning of algorithms.
Categorical and numerical parameters can be used as input variables, which are
specified in the \ROI file. A configuration file (\CONF) collects data related
to the \SPOT run (which is considered as a project) such as the prediction
model.  The reader is referred to the {\tt SpotGetOptions} help page, which
lists \SPOT's configuration parameters.

Parameters related to the algorithm or the optimization problem are stored in the \APD file.
\SPOT generates simple text files which are used as interfaces to the algorithm.

The sequential approach comprehends the following steps:
\begin{itemize}
  \item \init: generate an initial design
  \item \run: evaluate the algorithm
  \item \seq: generate new design points (meta model)
  \item \rep: statistical analysis and visualization, \EDA
\end{itemize}
Plugins for these steps are subject of on-going research.
\SPOT can also be run in a \meta mode to perform tuning over a set of problem
instances. 
Plugin development concentrates on combining predictions from several
regression models, integrating tools for multi objective optimization, and
performing meta \SPOT runs.

The \SPOT package contains several demos, which can be used as starting points
for setting up your own \SPOT project. Use 
\begin{Schunk}
\begin{Sinput}
> demo(spotDemo07RandomForestSann, ask=FALSE)
\end{Sinput}
\end{Schunk}
to start a demo which is related to the experimental setup
from Sect.~\ref{sec:tunespot}.

\section{Acknowledgements}
This work was supported by the Bundesministerium f\"ur Bildung und
Forschung (BMBF) under the grants FIWA (AIF FKZ 17N2309, "Ingenieurnachwuchs") 
and SOMA (AIF FKZ 17N1009, "Ingenieurnachwuchs")
and by the Cologne University of Applied Sciences under the research focus grant COSA. 

Many thanks go to members of the FIWA and SOMA research group.

\section{Appendix}
\subsection{\R Source Code for Starting \SANN}
\lstset{caption ={spotAlgStartSann.R}\label{lst:run1}}
\begin{lstlisting}[firstline=40]
spotAlgStartSann <- function(io.apdFileName, io.desFileName, io.resFileName){
	writeLines(paste("Loading design file data from::",  io.desFileName), con=stderr());
	source( io.apdFileName,local=TRUE)
	des <- read.table( io.desFileName, sep=" ", header = TRUE);	 
	pNames <- names(des);	
	config<-nrow(des);	
	for (k in 1:config){
		for (i in 1:des$REPEATS[k]){
			if (is.element("TEMP", pNames)){
				temp <- des$TEMP[k]
			}
			if (is.element("TMAX", pNames)){
				tmax <- round(des$TMAX[k])
			}
			conf <- k
			if (is.element("CONFIG", pNames)){
				conf <- des$CONFIG[k]
			}
			spotStep<-NA
			if (is.element("STEP", pNames)){
				spotStep <- des$STEP[k]
			}			
			seed <- des$SEED[k]+i-1	
			set.seed(seed)
			y <- optim(x0, spotFuncStartBraninSann, method="SANN",
					control=list(maxit=maxit, temp=temp, tmax=tmax, parscale=parscale))	
			res <- NULL
			res <- list(Y=y$value, TEMP=temp, TMAX=tmax, FUNCTION=f, DIM=n, SEED=seed, CONFIG=conf)
			if (is.element("STEP", pNames)){
				res=c(res,STEP=spotStep)
			} 
			res <-data.frame(res)
			colNames = TRUE
			if (file.exists(io.resFileName)){
				colNames = FALSE
			}
			write.table(res, file = io.resFileName, row.names = FALSE, 
					col.names = colNames, sep = " ", append = !colNames, quote = FALSE);			
			colNames = FALSE
		}
	}	
}
\end{lstlisting}

\subsection{\R Source Code for Combining Meta Models}
\lstset{caption={spotPredictRandomForestMlegp.R}\label{lst:rfmlegp1}}
\begin{lstlisting}[firstline=20]
	spotInstAndLoadPackages("mlegp")	
	xNames <- setdiff(names(rawB),"y")
	x <- rawB[,xNames]	
	y <- rawB$y	
	rf.fit <- randomForest(x, y)		
	rf.res <- predict(rf.fit,largeDesign)
	rf.largeDesign <- largeDesign[order(rf.res,decreasing=FALSE),]
	rf.s <- round(spotConfig$seq.design.new.size/2)
	mlegp.s <- spotConfig$seq.design.new.size - rf.s
	rf.largeDesign <- rf.largeDesign[1:rf.s,]
	if (mlegp.s> 0){
		mlegp.fit <- mlegp(x, y)		
		mlegp.res <- predict(mlegp.fit,largeDesign)
		mlegp.largeDesign <- largeDesign[order(rf.res,decreasing=FALSE),]
		mlegp.largeDesign <- largeDesign[1:mlegp.s,]
		return(rbind(rf.largeDesign,mlegp.largeDesign))
	}
	else{ return(rf.largeDesign)}
	writeLines("spotPredictRandomForestMlegp finished");
	return(largeDesign)
}
\end{lstlisting}

\subsection{\R Source Code for the Comparison of SANN Parameter Settings}
First, we will set the seed to obtain reproducible results. 
\begin{Schunk}
\begin{Sinput}
> set.seed(1)
\end{Sinput}
\end{Schunk}
Next, we will define the objective function.
\begin{Schunk}
\begin{Sinput}
> spotFunctionBranin <- function(x) {
+     x1 <- x[1]
+     x2 <- x[2]
+     (x2 - 5.1/(4 * pi^2) * (x1^2) + 5/pi * x1 - 6)^2 + 10 * (1 - 
+         1/(8 * pi)) * cos(x1) + 10
+ }
\end{Sinput}
\end{Schunk}
Then, the starting point for the optimization $x_0$ and the number of function evaluations $\maxit$ are defined:
\begin{Schunk}
\begin{Sinput}
> x0 <- c(10, 10)
> maxit <- 250
\end{Sinput}
\end{Schunk}
The parameters specified so far belong to the problem design. 
Now we have to consider parameters from the algorithm design, i.e., parameters that 
control the behavior of the SANN algorithm, namely \tmax and \temp:
\begin{Schunk}
\begin{Sinput}
> tmax <- 10
> temp <- 10
\end{Sinput}
\end{Schunk}
Finally, we can start the optimization algorithm (SANN):
\begin{Schunk}
\begin{Sinput}
> y1 <- NULL
> for (i in 1:10) {
+     set.seed(i)
+     y1 <- c(y1, optim(x0, spotFunctionBranin, method = "SANN", 
+         control = list(maxit = maxit, temp = temp,
+         tmax = tmax))$value)
+ }
> summary(y1)
\end{Sinput}
\begin{Soutput}
   Min. 1st Qu.  Median    Mean 3rd Qu.    Max. 
 0.3995  0.4037  0.4174  0.9716  0.6577  4.0670 
\end{Soutput}
\end{Schunk}
\begin{Schunk}
\begin{Sinput}
> temp <- 1.283295
> tmax <- 41
> y2 <- NULL
> for (i in 1:10) {
+     set.seed(i)
+     y2 <- c(y2, optim(x0, spotFunctionBranin, method = "SANN", 
+         control = list(maxit = maxit, temp = temp,
+         tmax = tmax))$value)
+ }
> summary(y2)
\end{Sinput}
\begin{Soutput}
   Min. 1st Qu.  Median    Mean 3rd Qu.    Max. 
 0.3981  0.3999  0.4007  0.4018  0.4035  0.4085 
\end{Soutput}
\end{Schunk}
\begin{Schunk}
\begin{Sinput}
> temp <- 0.1
> tmax <- 1
> y3 <- NULL
> for (i in 1:10) {
+     set.seed(i)
+     y3 <- c(y3, optim(x0, spotFunctionBranin, method = "SANN", 
+         control = list(maxit = maxit, temp = temp,
+         tmax = tmax))$value)
+ }
> summary(y3)
\end{Sinput}
\begin{Soutput}
   Min. 1st Qu.  Median    Mean 3rd Qu.    Max. 
 0.3980  0.3982  0.3984  0.3995  0.3989  0.4047 
\end{Soutput}
\end{Schunk}
%
%
\bibliographystyle{apalike}
\bibliography{arxivbartz}

\end{document}